\documentclass[12pt]{article}
\usepackage[a4paper, total={6in, 8in}]{geometry}
\usepackage{times}
\usepackage{authblk}
\usepackage{graphicx}
\usepackage{amssymb}
\usepackage{algorithm}
\usepackage{algorithmic}
\usepackage{amsmath}
\usepackage{subfigure}
\usepackage{makecell}
\usepackage{bbm}
\usepackage{color}
\usepackage{wrapfig}

\newcommand{\vect}[1]{{\boldsymbol{#1}}}
\newcommand{\intd}{\textrm{d} }
\newcommand{\E}{\mathbb{E} }
\newcommand{\R}{\mathbb{R} }
\newcommand{\KL}{D_{\textrm{KL}}}

\newcommand{\blue}[1]{{\color{black}#1}}

\begin{document}

%\begin{frontmatter}


\title{\Large Discovering Diverse Solutions in Deep Reinforcement Learning by Maximizing State-Action-Based Mutual Information }

 \author[1,2]{\normalsize Takayuki Osa}
 \affil[1]{\footnotesize Kyushu Institute of Technology, 
             Kita-kyushu,
             808-0135,
             Fukuoka,
             Japan}

 \affil[2]{RIKEN Center for Advanced Intelligence Project, 
             Chuo-ku,
             103-0027,
             Tokyo,
             Japan}

\author[2]{Voot Tangkaratt}

\author[2,3]{Masashi Sugiyama}

\affil[3]{The University of Tokyo, %Department and Organization
            Kashiwa,
            277-8561, 
            Chiba,
            Japan}
\date{}
        
 \maketitle

\begin{abstract}
Reinforcement learning algorithms are typically limited to learning a single solution for a specified task, even though diverse solutions often exist. 
Recent studies showed that learning a set of diverse solutions is beneficial because diversity enables robust few-shot adaptation. 
Although existing methods learn diverse solutions by using the mutual information as unsupervised rewards, 
such an approach often suffers from the bias of the gradient estimator induced by value function approximation.
In this study, we propose a novel method that can learn diverse solutions without suffering the bias problem. In our method, a policy conditioned on a continuous or discrete
latent variable is trained by directly maximizing the variational lower bound of the mutual information, 
instead of using the mutual information as unsupervised rewards as in previous studies.
Through extensive experiments on robot locomotion tasks, we demonstrate that the proposed method successfully learns an \emph{infinite} set of diverse solutions by learning continuous latent variables, 
which is more challenging than learning a finite number of solutions. 
Subsequently, we show that our method enables more effective few-shot adaptation compared with existing methods. 
\end{abstract}



%\begin{keyword}
%	Reinforcement learning \sep Robot learning \sep Representation learning



%\end{keyword}

%\end{frontmatter}


\section{Introduction}
\label{sec:intro}
Reinforcement learning~(RL) has achieved significant success in various fields, including robotic manipulation~\cite{Levine16b,Bodnar20}, board games~\cite{silver16}, and optimization of experimental setups~\cite{Sorokin20}.
In RL, multiple optimal policies may exist that elicit the optimal value function~\cite{Sutton18}.
For example, when learning a walking behavior at a certain speed, there are often an infinite number of walking styles to achieve the specified speed.
However, many existing RL methods are typically limited to obtaining only one of the optimal policies in a stochastic manner.

Recent studies investigated methods for learning diverse solutions and demonstrated their benefits. 
For instance, \cite{Kumar20} demonstrated that unseen tasks can be solved rapidly by adapting diverse solutions of seen tasks, and 
\cite{Eysenbach19} showed that  learned diverse skills can be used as low-level policies of hierarchical RL for the performance improvement. 
Existing methods for learning diverse behaviors often maximize the mutual information (MI) by treating MI as unsupervised rewards~\cite{Eysenbach19,Kumar20}.
However,  when treating MI as unsupervised rewards, we face the problem of function approximation which significantly affects the performance of actor-critic methods~\cite{Fujimoto18}.
To stabilize the process for learning diverse solutions, it is desirable to develop a strategy for maximizing MI in a more stable manner.

\begin{figure}
	\centering
	\includegraphics[width=1.0\columnwidth]{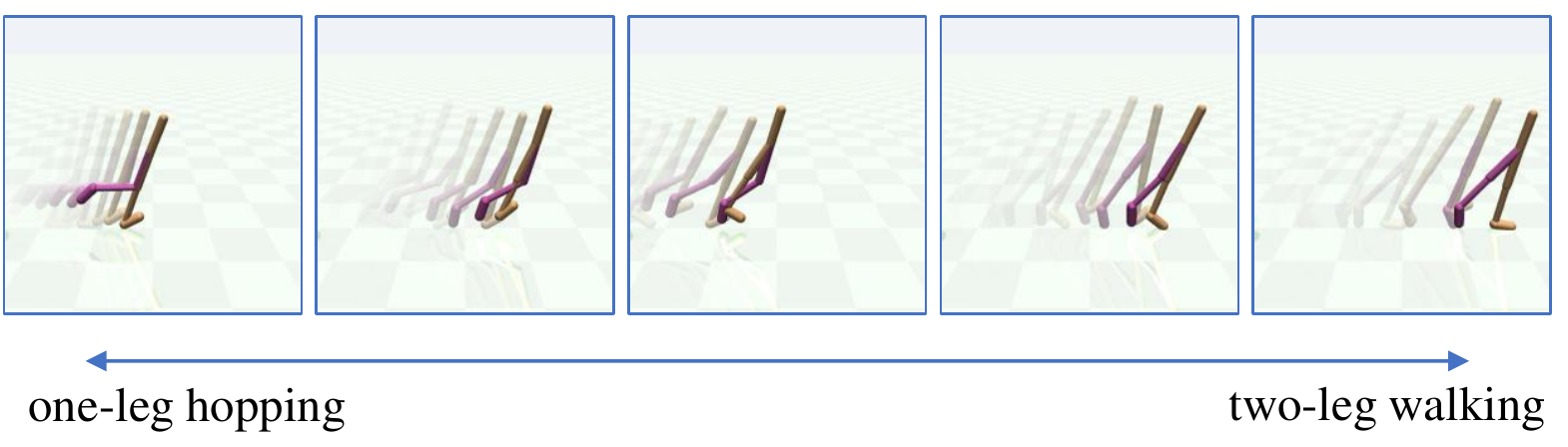}
	\caption{The proposed method successfully learned the continuous skill space that encodes diverse behaviors. In the leftmost figure, the agent is performing the one-leg hopping, while the agent is performing the two-leg walking with stretched knees in the rightmost figure. The walking style can be continuously changed by updating the value of the latent variable continuously.}
	\label{fig:intro}
	\vspace{1.0cm}
\end{figure}

The key contribution of our work is the proposal of a practical algorithm, \emph{latent-conditioned TD3~(LTD3)}, which can train a policy conditioned on a continuous or discrete latent variable and learn an \emph{infinite} number of solutions.
In our method, a policy conditioned on a latent variable is trained by directly maximizing a variational lower bound of MI, 
instead of treating MI as unsupervised rewards as in the previous studies~\cite{Eysenbach19,Kumar20,Sharma20}.
Our method enables us to avoid the bias of the gradient estimator induced by value function approximation and steadily maximize MI.
In addition, we also utilize a truncated importance sampling scheme for maximizing MI in a sample-efficient manner.  
Moreover, our method can be regarded as an approach that leverages separate critics for extrinsic and intrinsic rewards with different discount factors as in~\cite{Burda19, Badia20}.
We empirically investigated different strategies for maximizing MI and show that the our approach is sample-efficient and stable.
While previous studies~\cite{Eysenbach19,Kumar20,Parker-Holder20} often focused on learning a finite number of solutions, 
our experimental results demonstrate that our method can successfully learn a continuous skill space that encodes infinitely diverse behaviors, which is more challenging. $\Phi$
The effectiveness of our method in terms of few-shot robustness~\cite{Kumar20} was experimentally evaluated as shown in Figure~\ref{fig:intro}, and our method outperformed baseline methods.

\section{Background}
In this section, we describe definitions of terms related to an RL problem. Subsequently, we introduce the latent-conditioned policies for learning multiple solutions and describe the definition of few-shot robustness.
 
\subsection{Reinforcement Learning}
We consider an RL problem under a Markov decision process~(MDP) defined by a tuple $(\mathcal{S}, \mathcal{A}, \mathcal{P}, r  , \gamma, d)$ where $\mathcal{S}$ is the state space,  $\mathcal{A}$ is the action space,  $\mathcal{P}(\vect{s}_{t+1}|\vect{s}_t, \vect{a}_t)$ is the transition probability density,  $r(\vect{s}, \vect{a})$ is the reward function, $\gamma$ is the discount factor, and $d(\vect{s}_0)$ is the probability density of the initial state.
A policy $\pi(\vect{a}|\vect{s}): \mathcal{S} \times \mathcal{A} \mapsto \R^+$ is defined as the conditional probability density over actions given states.
The goal of RL is to identify a policy that maximizes the expected cumulative discounted reward $\E[R_0|\pi] $ where $R_t$ is a return, which is given by $R_t = \sum_{k=t}^{T} \gamma^{k-t}r(\vect{s}_k, \vect{a}_k)$.

\subsection{Latent-Conditioned Policies for Learning Multiple Solutions}
In this study, we focus on tasks in which multiple, possibly infinitely many policies achieve the maximal expected return.
This scenario can be observed in many practical tasks, particularly in robotics~\cite{Toussaint18,Orthey20,osa20,Osa21}.
For example, when learning the walking behavior of a robot, many walking motions can achieve the same speed, as shown in the top row of Figure~\ref{fig:intro}.

To model diverse behaviors, we consider a policy conditioned on a latent variable as in previous studies~\cite{Kumar20,Eysenbach19,Sharma20}.
Herein, we introduce latent variable $\vect{z} \in \mathcal{Z}$, where $\mathcal{Z}$ is a set of possible values of $\vect{z}$ that can be discrete or continuous.
We consider a policy of the form
\begin{align}
	\pi(\vect{a}|\vect{s}) = \int \pi(\vect{a}|\vect{s}, \vect{z}) p(\vect{z}) \intd \vect{z},
	\label{eq:latent_policy}
\end{align}
where $\vect{z}$ is the latent variable and $p(\vect{z})$ is the prior distribution of $\vect{z}$.
The conditional probability density of observing a trajectory $\vect{\tau}$ given the latent variable $\vect{z}$ is expressed as
\begin{align}
	p(\vect{\tau} | \vect{z}) = d(\vect{s}_0) \prod_{t=0}^{T} \pi(\vect{a}_t | \vect{s}_t, \vect{z}) \mathcal{P}(\vect{s}_{t+1} | \vect{s}_t, \vect{a}_t).
\end{align}
In our setting, $\vect{z}$ is sampled at the start of a trajectory and does not change until the end of the trajectory as in previous studies~\cite{Kumar20,Eysenbach19}.
We also introduce the Q-function conditioned on the latent variable, $Q^\pi(\vect{s}, \vect{a}, \vect{z})$, which satisfies the Bellman equation as
\begin{align}
	Q^\pi(\vect{s}, \vect{a}, \vect{z}) = r(s,a) + \gamma \E_{\vect{s}' \sim p(\vect{s}'|\vect{s}, \vect{a}), \vect{a} \sim \pi(\vect{a}'|\vect{s}',\vect{z})}[Q^\pi(\vect{s}',\vect{a}',\vect{z})].
\end{align}
We refer to $Q^\pi(\vect{s}, \vect{a}, \vect{z})$ as the latent-conditioned action-value function, and it represents the expected return when following a policy for the latent variable.
In this study, we aim to learn a policy that generates various trajectories by changing the value of latent variable $\vect{z}$.

\subsection{Few-Shot Robustness}
Researchers have demonstrated the usefulness of learning diverse solution in terms of few-shot robustness~\cite{Kumar20} and hierarchical RL~\cite{Eysenbach19}. 
In this paper, we focus on utilizing diverse solution to improve few-shot robustness of RL agent. Specifically, we follow the protocol for evaluating few-shot robustness described in~\cite{Kumar20}.
In the training phase, a policy is trained in a training MDP given by  $\mathcal{M}=(\mathcal{S}, \mathcal{A}, \mathcal{P}, r  , \gamma, d)$. Subsequently, the policy is tested in a test MDP with $\mathcal{M}'=(\mathcal{S}, \mathcal{A}, \mathcal{P}', r'  , \gamma, d')$, in which the state and action space are identical but the transition probability density, initial state density and reward function differ from those of the training MDP.
In the test MDP, the policy should be adapted after a limited number of trials and then the best policy is selected.
The number of episodes in which the agent is allowed to interact with the test MDP is referred to as the \textit{budget}.
The performance of the policy after adaptation is evaluated under the test MDP.
In this study, we compared the few-shot robustness of policies trained using our method against those trained using existing methods.

\section{Learning Latent Representations of Policies}
In this section, we formulate the problem of learning diverse solutions in RL. Subsequently, we describe our method to maximize the MI based on a variational lower bound.

\subsection{Problem Formulation}
Our goal is to obtain a latent-conditioned policy that encodes diverse solutions for a given MDP.
Although we can train the latent-conditioned policy by maximizing the expected return $\E[R|\pi]$, 
such a policy may disregard the latent variable and would not represent diverse solutions after training.
Hence, we propose maximizing the MI between the latent and state-action variables.
Specifically, to encode diverse behaviors in the latent-conditioned policy, 
we aim to solve the problem formulated as
\begin{align}
	\max_{\pi} \left( \E[ \blue{R_0}|\pi] + \alpha I_\pi(\vect{s}, \vect{a}; \vect{z}) \right),
	\label{eq:problem}
\end{align}
where  $\E[R|\pi]$ is the expected return of $\pi$, $\alpha$ is a constant that balances the two terms, and $I_\pi(\vect{s}, \vect{a}; \vect{z})$ is the MI between a state action pair $(\vect{s}, \vect{a})$ and latent variable $\vect{z}$ under policy $\pi$.
In our framework, we maximize the MI $I_{\pi}(\vect{s}, \vect{a}; \vect{z})$ defined as
\begin{align}
	I_{\pi}(\vect{s}, \vect{a}; \vect{z}) =  \iiint & p_{\pi}(\vect{s}, \vect{a}, \vect{z}) \log \left( \frac{p_{\pi}(\vect{s}, \vect{a}, \vect{z})}{p_{\pi}(\vect{s}, \vect{a}) p(\vect{z})} \right) \intd \vect{z} \intd \vect{a} \intd \vect{s}, 
\end{align}
where 
\begin{align}
p_{\pi}(\vect{s}, \vect{a}, \vect{z}) & = \beta(\vect{s}, \vect{z}) \pi(\vect{a}|\vect{s}, \vect{z}) \nonumber \\
p_{\pi}(\vect{s}, \vect{a}) & = \int p_{\pi}(\vect{s}, \vect{a}, \vect{z}) \intd \vect{z}, \nonumber
\end{align}
and $\beta(\vect{s}, \vect{z})$ is the joint density of states and latent variables induced by the behavior policy.
MI $I_{\pi}(\vect{s}, \vect{a}; \vect{z})$ is given by 
\begin{align}
	I_{\pi}(\vect{s}, \vect{a}; \vect{z}) = H(\vect{s}, \vect{a}) - H(\vect{a}|\vect{s}, \vect{z}) - H(\vect{s}| \vect{z}),
	\label{eq:I_decomp} 
\end{align}
where $H(\vect{s}, \vect{a})$ is the marginal entropy of $\vect{s}$ and $\vect{a}$, $H(\vect{a}|\vect{s}, \vect{z})$ is the conditional entropy of  $\vect{a}$ conditioned on $\vect{s}$ and $\vect{z}$, and they are given by
\begin{align}
	H(\vect{s}, \vect{a}) & = \iint p(\vect{s}, \vect{a}) \log p(\vect{s}, \vect{a}) \intd \vect{a}\intd \vect{s}, \\
	H(\vect{a}|\vect{s}, \vect{z}) & = \iiint p(\vect{s}, \vect{a}, \vect{z}) \log p(\vect{a} | \vect{s}, \vect{z}) \intd \vect{a} \intd \vect{s} \intd \vect{z},
\end{align}
and $p(\vect{s},\vect{a})$ is the marginal distribution of $\vect{s}$ and $\vect{a}$.
Therefore, when MI  $I_{\pi}(\vect{s}, \vect{a}; \vect{z})$ is maximized, $	H(\vect{s}, \vect{a})$ is increased and thus, a wide variety of actions will be covered by the trained model.
At the same time, $H(\vect{a}|\vect{s}, \vect{z})$ is decreased, and the conditional distribution of action $\vect{a}$ conditioned on $\vect{s}$ and $\vect{z}$ will be more deterministic.
Thus, we can obtain diverse behavior by maximizing  $I_{\pi}(\vect{s}, \vect{a}; \vect{z})$, and the behavior can be controlled by setting the value of $\vect{z}$.

\subsection{Information Maximization via Variational Lower Bound}
\label{sec:Infomax_bound}
In practice, directly computing $I_{\pi}(\vect{s}, \vect{a}; \vect{z})$ is challenging because it requires the density estimations of $ p_{\pi}(\vect{s}, \vect{a})$ and $p_{\pi}(\vect{s}, \vect{a}, \vect{z})$.
Hence, we use a variational lower bound of MI as the surrogate objective.
The variational lower bound of the MI between the state-action pair $(\vect{s}, \vect{a})$ and the latent variable $\vect{z}$ can be derived as follows:
\begin{align}
	I_{\pi}(\vect{s}, \vect{a}; \vect{z}) & = H(\vect{z}) - H(\vect{z} | \vect{s}, \vect{a} ) = \E_{(\vect{s}, \vect{a}, \vect{z}) \sim p_{\pi}} \left[ \log p(\vect{z} | \vect{s}, \vect{a}) \right] +  H(\vect{z}) \nonumber \\
	& = \E_{(\vect{s}, \vect{a}) \sim \beta(\vect{s}, \vect{a})} \left[ \KL\left( p(\vect{z} | \vect{s}, \vect{a})  || q_{\vect{\phi}}(\vect{z} | \vect{s}, \vect{a}) \right)\right]  + \E_{(\vect{s}, \vect{a}, \vect{z}) \sim p_{\pi}} \left[ \log  q_{\vect{\phi}}(\vect{z} | \vect{s}, \vect{a}) \right] +  H(\vect{z}) \nonumber \\
	& \geq  \E_{(\vect{s}, \vect{a}, \vect{z}) \sim p_{\pi}} \left[ \log  q_{\vect{\phi}}(\vect{z} | \vect{s}, \vect{a}) \right] +  H(\vect{z})
	\label{eq:info_bound}
\end{align}
Herein, the density model $q_{\vect{\phi}}(\vect{z} | \vect{s}, \vect{a})$ parameterized by a vector $\vect{\phi}$ is introduced to approximate the posterior density $ p(\vect{z} | \vect{s}, \vect{a})$, and 
$p_{\pi}(\vect{s}, \vect{a}, \vect{z})$ is the joint density of state $\vect{s}$, action $\vect{a}$, and latent variable $\vect{z}$ stored in replay buffer $\mathcal{D}$.
$ \KL\left( p(\vect{z} | \vect{s}, \vect{a})  || q_{\vect{\phi}}(\vect{z} | \vect{s}, \vect{a}) \right)$ is the Kullback-Leibler divergence between $p(\vect{z} | \vect{s}, \vect{a})$ and $ q_{\vect{\phi}}(\vect{z} | \vect{s}, \vect{a})$.
We regarded entropy $H(\vect{z})$ as a constant with respect to the policy parameter, and we set $p(\vect{z})$ to be uniform to achieve a high $H(\vect{z})$.
Although the decomposition of MI in the first line of \eqref{eq:info_bound} is different from \eqref{eq:I_decomp}, the intuition described in the previous section still applies.

It was demonstrated that this lower bound is tight~\cite{Jordan99,Barber03}. 
Specifically, the surrogate on the right-hand side of \eqref{eq:info_bound} is equivalent to the MI $I_{\pi}(\vect{s},\vect{a}; \vect{z})$ for the optimal variational distribution 
\[q^*_{\vect{\phi}} = \blue{\arg} \max_{q_{\vect{\phi}}} \E[\log q_{\vect{\phi}}( \vect{z}| \vect{s}, \vect{a})] + H(\vect{z}). \]
Therefore, we may replace the MI in \eqref{eq:problem} with $\max_{q_\phi} \E[\log q_\vect{\phi}( \vect{z}| \vect{s}, \vect{a})] + H(\vect{z})$ to yield the following optimization problem:
\begin{align}
	\max_{\pi,\vect{\phi}} \left( \E[R|\pi] + \E[\log q_\vect{\phi}( \vect{z}| \vect{s}, \vect{a})] \right).
\end{align}

\subsection{Importance Sampling in Information Maximization for Learning Diverse Solutions }
The previous study \cite{Kumar20} proposed maximizing the MI based on samples with high returns in the replay buffer by maxmizing the following term
\[
 \sum_{(\vect{s}, \vect{a}, \vect{z}) \in \mathcal{D}} \mathbbm{1}_{R(\pi) > R(\pi^*) - \epsilon }(\vect{s}, \vect{a}, \vect{z}) \cdot r_{\textrm{info}},
\]
where $R(\pi^*)$ is the return obtained by the optimal policy, $\epsilon$ is a constant, $r_{\textrm{info}}$ is the reward related to the mutual information, and $\mathbbm{1}_{R(\pi) > R(\pi^*) - \epsilon }(\vect{s}, \vect{a}, \vect{z})$ is the indicator function. 
$\mathbbm{1}_{R(\pi^*) > R(\pi) - \epsilon }(\vect{s}, \vect{a}, \vect{z})$ is 1 when the return of the episode in which $(\vect{s}, \vect{a}, \vect{z})$ is observed is greater than $R(\pi^*) - \epsilon$, otherwise 0.
Intuitively, samples from episodes with near-optimal returns are selected from the replay buffer, and rewards for maximizing MI is given to those samples. 
However, the method requires running an off-the-shelf RL method to approximate the return achieved by the optimal policy; therefore, it is not sample-efficient.
In our method, we avoid this by applying the importance weight.
To approximate the density of $\vect{s}, \vect{a}$, and $\vect{z}$ induced by the optimal policy, we consider a Boltzmann distribution
\[
f(\vect{s}, \vect{a}, \vect{z}) = \frac{\exp(Q(\vect{s}, \vect{a}, \vect{z}))}{Z},
\]
where $Z = \iiint \exp(Q(\vect{s}, \vect{a}, \vect{z})) \intd \vect{s}\intd \vect{a}\intd \vect{z}$, and $Q(\vect{s},\vect{a}, \vect{z})$ is the latent-conditioned action-value function.
In our method, we approximate the expectation with respect to samples selected from the replay buffer by following $f(\vect{s}, \vect{a}, \vect{z})$, instead of selecting the samples based on the indicator function $\mathbbm{1}_{R(\pi) > R(\pi^*) - \epsilon }(\vect{s}, \vect{a}, \vect{z})$.
For this purpose, we use the importance weight 
\begin{align}
	W = \frac{f(\vect{s}, \vect{a}, \vect{z})}{p_{\textrm{prop}}(\vect{s}, \vect{a}, \vect{z})} = \frac{\exp(Q(\vect{s}, \vect{a}, \vect{z}))}{Z \cdot p_{\textrm{prop}}(\vect{s}, \vect{a}, \vect{z})},
\end{align}
where $p_{\textrm{prop}}(\vect{s}, \vect{a}, \vect{z})$ is a proposal distribution, and we use the uniform distribution $U(\vect{s}, \vect{a}, \vect{z})$ as $p_{\textrm{prop}}(\vect{s}, \vect{a}, \vect{z})$ in our implementation.
Using the importance weight, we compute the expectation with respect to the density induced by selecting samples from the replay buffer by following  $f(\vect{s}, \vect{a}, \vect{z})$ as follows:
\begin{align}
	\mathcal{J}_{\textrm{info}}(\pi, \vect{\phi}) & = \E_{(\vect{s}, \vect{a}, \vect{z}) \sim f(\mathcal{D})} \left[ \log  q_{\vect{\phi}}(\vect{z} | \vect{s}, \vect{a}) \right] \nonumber \\
	& =  \E_{(\vect{s}, \vect{a}, \vect{z}) \sim U(\mathcal{D})} \left[ W(\vect{s}, \vect{a}, \vect{z}) \log  q_{\vect{\phi}}(\vect{z} | \vect{s}, \vect{a}) \right].
	\label{eq:info_priority}
\end{align}
The objective function presented in \eqref{eq:info_priority} indicates that a batch of data $(\vect{s}, \vect{a}, \vect{z})$ is sampled uniformly from the data stored in the replay buffer, and the samples used in the policy update are not correlated. However, this does not mean that the values of the latent variables must be uncorrelated during the data collection process.
In practice, we normalize the importance weight using a mini-batch  $\mathcal{D}_{\textrm{batch}}$ selected from the replay buffer and compute the importance weight:
\begin{align}
	\hat{W} =  \frac{\exp(Q(\vect{s}, \vect{a}, \vect{z}))}{\sum_{(\bar{\vect{s}}, \bar{\vect{a}},  \bar{\vect{z}}) \in \mathcal{D}_{\textrm{batch}}}\exp\big(Q(\bar{\vect{s}}, \bar{\vect{a}},  \bar{\vect{z}})\big)}.
\end{align}
Here, the partition function $Z$ is canceled when normalizing the importance weight.
Using this importance weight, we prioritize samples with high action values when optimizing the MI.
However, the importance-sampling technique induces high variances. 
Hence, we propose using a truncated importance weight $\tilde{W}$ which is defined as 
\begin{align}
	\tilde{W} = 
	\max\left(1 - c_{\textrm{clip}}, \min\left(\hat{W}, 1 + c_{\textrm{clip}}\right)\right),
	\label{eq:clip}
\end{align}
where $c_{\textrm{clip}} > 0$ is a hyperparameter. 
In \eqref{eq:clip}, $W$ smaller than $1 - c_{\textrm{clip}}$ is rounded up to $ 1 - c_{\textrm{clip}}$ and $W$ larger than $ 1 + c_{\textrm{clip}}$  is rounded down to $ 1 + c_{\textrm{clip}}$.
Similar importance weight clipping is often employed to stabilize the learning~\cite{Munos16,Han19} 
The effect of the parameter $c_{\textrm{clip}}$ is discussed in the experiment section.

Based on the discussion above, we propose training a latent-conditioned policy by solving
\begin{align}
	\max_{\pi,\phi} \left( \E[R|\pi] + \mathcal{J}_{\textrm{info}}(\pi,\vect{\phi}) \right),
\end{align}
where $\mathcal{J}_{\textrm{info}}(\pi,\vect{\phi})$ is the importance-weighted lower bound in \eqref{eq:info_priority}. 
$\E[R|\pi]$ can be maximized using a gradient-based algorithm such as the policy gradient algorithm~\cite{Williams92}.
In the next section, we propose learning the latent-conditioned policy by extending a state-of-the-art actor-critic algorithm, namely the twin-delayed deep deterministic policy gradient (TD3) \cite{Fujimoto18}.

\section{Practical Algorithms for Training Latent-Conditioned Policy}
In this section, we present a practical algorithm for learning a latent-conditioned policy based on TD3~\cite{Fujimoto18}, which is an actor-critic algorithm that learns a parameterized policy and parameterized Q-function.
To approximate $Q^\pi(\vect{s}, \vect{a}, \vect{z})$, we employ a differentiable function approximator $Q_{\vect{w}}$ parameterized by a vector $\vect{w}$, which is hereinafter referred to as the critic network.

We consider a deterministic policy as in deep deterministic policy gradient (DDPG)~\cite{Lillicrap16} and TD3~\cite{Fujimoto18}. Therefore, $\pi(\vect{a}|\vect{s}, \vect{z})$ is a Dirac-delta function that satisfies
\begin{align}
	\int Q(\vect{s}, \vect{a}, \vect{z})\pi(\vect{a}|\vect{s}, \vect{z}) \intd \vect{a} =  Q(\vect{s}, \vect{\mu}_{\vect{\theta}}(\vect{s}, \vect{z}), \vect{z}),
\end{align}
where $\vect{\mu}_{\vect{\theta}}(\vect{s}, \vect{z}): \mathcal{S} \times \mathcal{Z} \mapsto \mathcal{A}$ is a model parameterized by a vector $\vect{\theta}$ that determines action $\vect{a}$ for a specified state $\vect{s}$ in a deterministic manner.
As in TD3, we employ two critic networks to stabilize the training, and they are trained by minimizing the mean squared error,
\begin{align}
	\mathcal{L}_{\textrm{critic}}(\vect{w}_i) = \E_{(\vect{s}, \vect{a}, \vect{s}', r, \vect{z}) \sim \blue{U(\mathcal{D})}} \left[ \left(Q_{\vect{w}_i}(\vect{s}, \vect{a}, \vect{z}) - y \right)^2 \right],
	\label{eq:q_loss}
\end{align}
for $i= 1,2$. Given $(\vect{s}, \vect{a}, \vect{s}', \vect{z}, r)$, the target Q-value $y$ is computed as
\begin{align}
	y = r + \gamma \min_{i=1,2} Q_{\vect{w}'_i} \left( \vect{s}', \vect{a}', \vect{z}) \right) \ \ \textrm{where} \  \
	\vect{a}' = \vect{\mu}_{\vect{\theta}}(\vect{s}, \vect{z}) + \vect{\epsilon}, \  \vect{\epsilon} \sim \mathcal{N}(0, \vect{\sigma}),
	\label{eq:ltd3_target}
\end{align}
and $\vect{\sigma}$ is the exploration noise variance.
Herein, we use the same value of $\vect{z}$ as $\vect{z}$ to compute the next action $\vect{a}'$. 
Therefore, the resulting $Q_{\vect{w}_i}(\vect{s}, \vect{a}, \vect{z})$ approximates the expected return when following $\pi$ with the value of $\vect{z}$ fixed during an episode.
The policy $\pi(\vect{a}|\vect{s}, \vect{z})$ is trained by maximizing 
\begin{align}
	\mathcal{J}(\vect{\theta}, \vect{\phi}) = \mathcal{J}_{Q}(\vect{\theta}) + \mathcal{J}_{\textrm{info}}(\vect{\theta}, \vect{\phi}),
\label{eq:objective}
\end{align}
where $\mathcal{J}_{Q}(\vect{\theta})$ is the term for maximizing the expected return based on the estimated Q-function, expressed as
\begin{align}
	\mathcal{J}_{Q}(\vect{\theta}) =  \E_{(\vect{s}, \vect{z}) \sim \blue{\beta}} \left[
	Q_{\vect{w}_1}(\vect{s}, \vect{\mu}_{\vect{\theta}}(\vect{s}, \vect{z} ), \vect{z} ) \right].
	\label{eq:actor_q_loss_ltd3}
\end{align}
The term $\mathcal{J}_{\textrm{info}}(\vect{\theta}, \vect{\phi})$ is expressed as 
\begin{align}
	\mathcal{J}_{\textrm{info}}(\vect{\theta}, \vect{\phi}) = \E_{(\vect{s}, \vect{z}) \sim \beta} \left[ \tilde{W} \log  q_{\vect{\phi}}(\vect{z} | \vect{s}, \vect{\mu_{\vect{\theta}}}(\vect{s}, \vect{z})) \right].
	\label{eq:info_loss_ltd3}
\end{align}
Herein, we refer to our TD3-based algorithm as Latent-conditioned TD3 (LTD3).

\begin{figure}
	\begin{minipage}{0.37\textwidth}
		\begin{figure}[H]
			\subfigure{\includegraphics[width=\columnwidth]{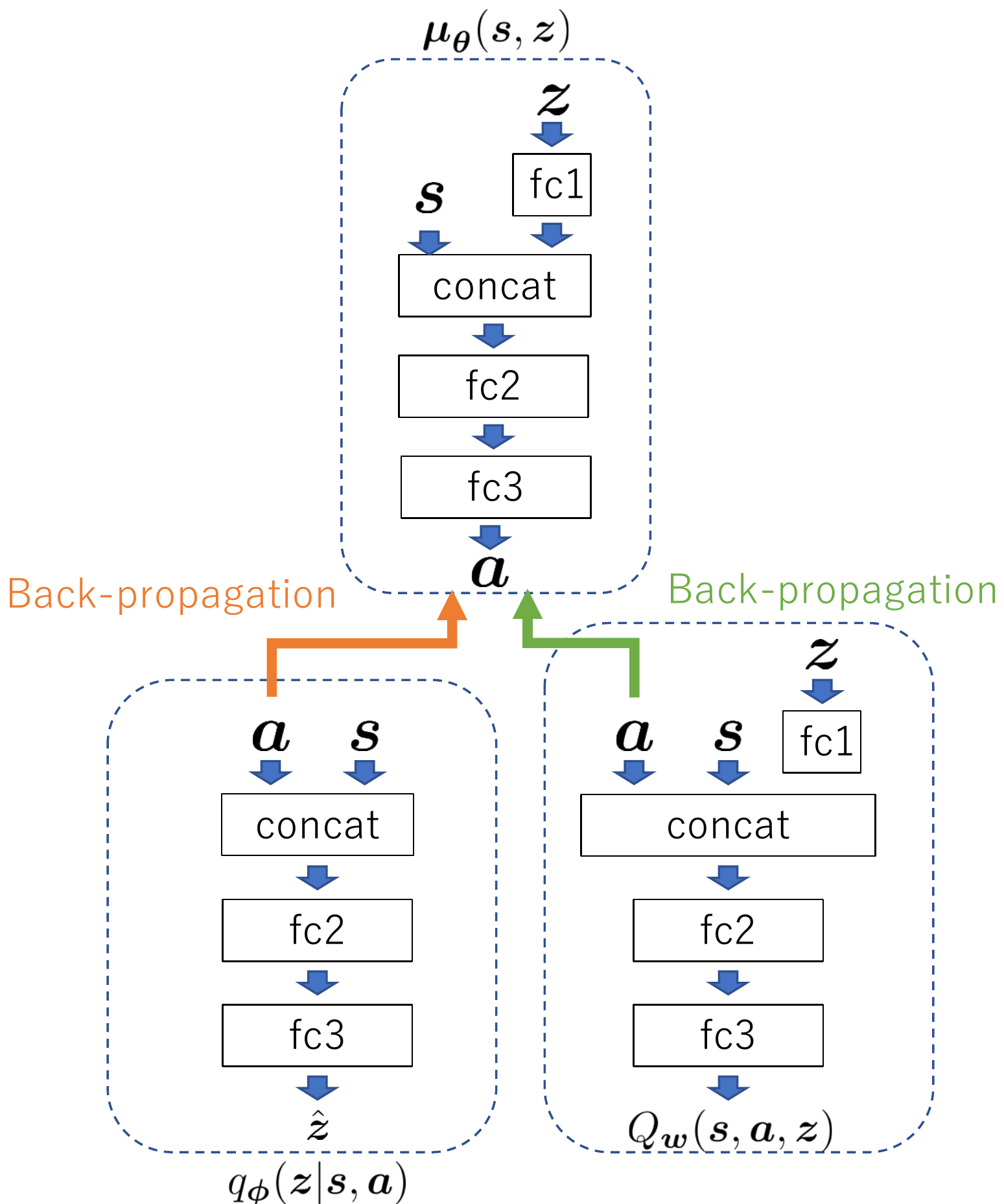}}
			\caption{Network architecture used in our implementation of LTD3.}
			\label{fig:network}
		\end{figure}
	\end{minipage}
	\hfill
	\begin{minipage}{0.6\textwidth}
		\begin{algorithm}[H]
			\caption{Latent-Conditioned TD3}
			\label{alg:lac}
			\begin{algorithmic}
				\STATE {\bfseries Input:} Dimension of latent variable $\vect{z}$
				\STATE Initialize the parameters $\vect{w}$, $\vect{\theta}$ and $\vect{\phi}$
				\STATE Initialize the replay buffer $\mathcal{D}$
				%   \REPEAT
				\FOR{each episode}
				\STATE Sample latent variable $\vect{z} \sim p(\vect{z})$
				\FOR{$t=0$ {\bfseries to} $T$}
				\STATE  Select action with exploration noise
				%\\ $\vect{a}= \vect{\mu}_{\vect{\theta}}(\vect{s}, \vect{z}) + \vect{\epsilon}$, where $\vect{\epsilon} \sim \mathcal{N}(0, \vect{\sigma})$
				\STATE Observe reward $r$ and new state $\vect{s}'$ 
				\STATE Store tuple $(\vect{s}, \vect{a}, \vect{s}', r, \vect{z})$ in $\mathcal{D}$ 
				\STATE Sample mini-batch from $\mathcal{D}$
				\STATE Update the critics by minimizing $\mathcal{L}_{\textrm{critic}}(\vect{w}_i)$ in \eqref{eq:q_loss} 
				\IF{$t$ mod $d_{\textrm{atr}}$}
				\STATE Update the actor by maximizing $\mathcal{J}_{Q}(\vect{\theta})$ in \eqref{eq:actor_q_loss_ltd3} 
				\ENDIF
				\IF{$t$ mod $d_{\textrm{info}}$}
				\STATE Update $\vect{\phi}$ and $\vect{\theta}$ by maximizing $\mathcal{J}_{\textrm{info}}(\vect{\theta})$ in \eqref{eq:info_loss_ltd3}
				\ENDIF
				\ENDFOR
				\ENDFOR
				%   \UNTIL{$noChange$ is $true$}
			\end{algorithmic}
		\end{algorithm}
	\end{minipage}
\end{figure}

The proposed algorithm is summarized in Algorithm~\ref{alg:lac}.
The network architecture used in the LTD3 implementation is shown in Figure~\ref{fig:network}.
In the network that models the latent-conditioned policy, the latent variable is input to a fully connected layer and then concatenated with the state vector.
In the implementation, the conditional density $q_{\vect{\phi}}(\vect{z} | \vect{s}, \vect{a})$ is modeled by a neural network.
We assume that the latent variable $\vect{z}$ is given by $\vect{z} = [\vect{z}_{\textrm{cont}}, \vect{z}_{\textrm{disc}}]$ where $\vect{z}_{\textrm{cont}}$ and $\vect{z}_{\textrm{disc}}$ are continuous and discrete variables, respectively.
We model $p(\vect{z}_{\textrm{cont}}| \vect{s}, \vect{a})$ as a factored Gaussian.
For the categorical latent variable $\vect{z}_{\textrm{disc}}$, the categorical distribution $q_{\vect{\phi}}(\vect{z}_{\textrm{disc}} | \vect{s}, \vect{a})$ is modeled with a softmax layer.
Whereas existing methods~\cite{Kumar20,Sharma20} regard the MI term as unsupervised rewards,
we directly maximize the information term via back-propagation in our method.    
In our framework, the gradient of the information term is back-propagated from the posteritor approximator to the policy network.

\section{Rationale of the Proposed Method}
\begin{figure}[tb]
	\centering
	\subfigure[In this MDP, two actions, $a_1$ and $a_2$ are allowed in each state. Reward $r=1$ is given when states $S_1$ or $S_4$ is visited.]{\includegraphics[width=0.35\columnwidth]{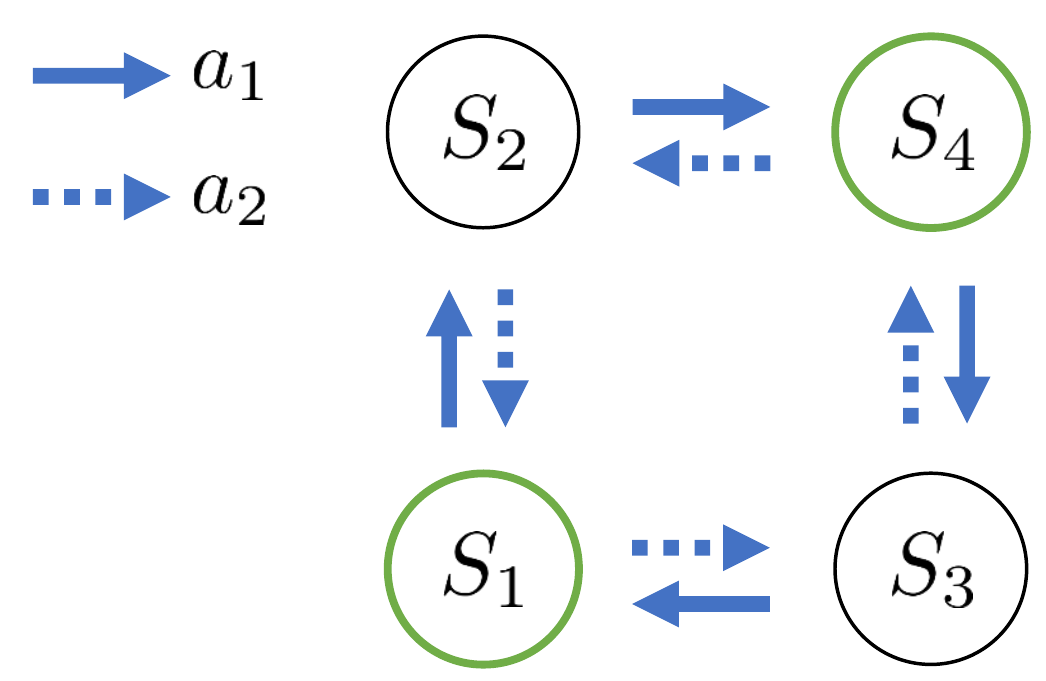}}
	\hfill
	\subfigure[Optimal Policy $\pi_1$.]{\includegraphics[width=0.25\columnwidth]{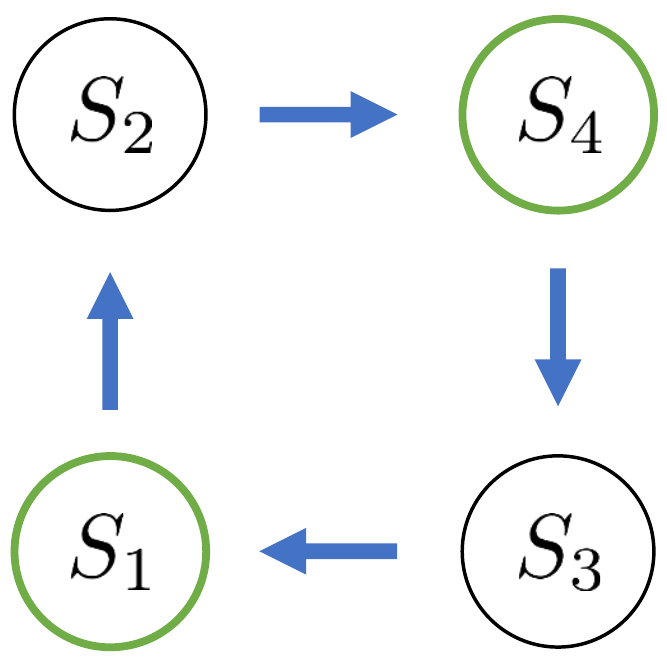}}
	\hfill
	\subfigure[Another optimal policy $\pi_2$, which is different from policy $\pi_1$.]{\includegraphics[width=0.25\columnwidth]{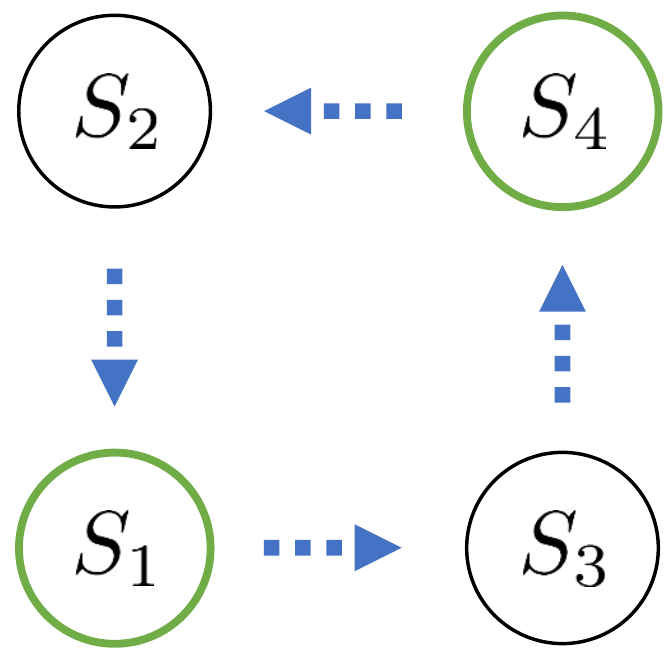}}
	\caption{Example of an MDP where maximizing $I(\vect{s}, \vect{a};\vect{z})$ can encourage more diversity than maximizing $I(\vect{s}; \vect{z})$. As all the states are equally visited by $\pi_1$ and $\pi_2$, $q(\vect{z}|\vect{s})$ cannot quantify the difference between $\pi_1$ and $\pi_2$. In contrast, $q(\vect{z}|\vect{s}, \vect{a})$ can quantify the difference between $\pi_1$ and $\pi_2$ by considering the corresponding action. }
	\vspace{0.7cm}
	\label{fig:example}
\end{figure}

In existing studies, the MI between the state and latent variable $I(\vect{s};\vect{z})$ was typically maximized~\cite{Eysenbach19,Kumar20}, which encourages different skills to visit different states. 
This strategy can encourage diversity in behaviors; 
however, maximizing state-only mutual information $I(\vect{s};\vect{z})$ may not be sufficient to encourage diversity of behaviors.
In Figure~\ref{fig:example}, we showcase an MDP where maximizing $I(\vect{s}, \vect{a};\vect{z})$ can encourage more diversity than maximizing $I(\vect{s}; \vect{z})$.
As shown in Figure~\ref{fig:example}(a), two actions, $a_1$ and $a_2$ are allowed in each state in the MDP, and reward $r=1$ is given when state $S_1$ or $S_4$ are visited.
different optimal policies for this MDP are shown in Figures~\ref{fig:example}(b) and (c). 
In practice, $I(\vect{s};\vect{z})$ is maximized by estimating the posterior distribution $q(\vect{z}|\vect{s})$, which predicts the value of the latent variable for a given state.
However, as all states are equally visited by $\pi_1$ and $\pi_2$, $q(\vect{z}|\vect{s})$ cannot quantify the difference between $\pi_1$ and $\pi_2$.
Thus, maximizing $I(\vect{s};\vect{z})$ cannot encourage the diversity of policies such as $\pi_1$ and $\pi_2$.
	
In contrast, we maximize $I(\vect{s}, \vect{a}; \vect{z})$ to encode the diversity of both states and actions into the latent variable. 
The action-state mutual information $I(\vect{s}, \vect{a};\vect{z})$ is maximized by estimating the posterior distribution $q(\vect{z}|\vect{s}, \vect{a})$, not $q(\vect{z}|\vect{s})$. 
Unlike $q(\vect{z}|\vect{s})$, the posterior distribution $q(\vect{z}|\vect{s}, \vect{a})$ can quantify the difference between $\pi_1$ and $\pi_2$ by considering the corresponding action.
Thus, maximizing $I(\vect{s}, \vect{a};\vect{z})$  can encourage diversity in policies such as $\pi_1$ and $\pi_2$.
This example indicates that there are situations where maximizing $I(\vect{s}; \vect{z})$ is not sufficient and maximizing $I(\vect{s}, \vect{a};\vect{z})$ is more suitable.
%Our experimental results in Section~\ref{sec:exp} indicate that maximizing $I(\vect{s}, \vect{a};\vect{z})$ result in more diverse and stable solutions for locomotion tasks compared with maximizing $I(\vect{s};\vect{z})$.

Although we discussed the benefit of maximizing $I(\vect{s}, \vect{a};\vect{z})$, there are tasks where maximizing $I(\vect{s};\vect{z})$ is sufficient. 
For example, previous studies have demonstrated that an agent can learn to walk in diverse directions by maximizing $I(\vect{s};\vect{z})$. 
When training an agent to walk in diverse directions, the most important information is the agent destination, which is included in state $\vect{s}$. 
In such cases, maximizing $I(\vect{s};\vect{z})$ can be sufficient because action $\vect{a}$ is not necessary to encourage the diversity of the walking directions.

In previous studies~\cite{Kumar20, Sharma20}, Soft Actor Critic~(SAC)\cite{Haarnoja18} was used as a base RL algorithm.
Although SAC is a state-of-the-art actor critic method as well as TD3, we did not use SAC as a main base algorithm; instead, we used the TD3-based implementation in our study because of the following reason.
If we use SAC as a base algorithm in our framework,  $H(\vect{a}|\vect{s}, \vect{z})$ will be maximized because of the entropy regularization in SAC.
In our framework, we maximize MI $I(\vect{s}, \vect{a}; \vect{z}) = H(\vect{s}, \vect{a}) - H(\vect{a}|\vect{s}, \vect{z}) -  H(\vect{s} |\vect{z})$.
Therefore, maximization of $H(\vect{a}|\vect{s}, \vect{z})$ can conflict with that of $I(\vect{s}, \vect{a}; \vect{z})$.
In this sense, it is reasonable to use a deterministic policy for $\pi(\vect{a}|\vect{s}, \vect{z})$, which can be viewed as the limit of the normal distribution $\mathcal{N}(\mu, \sigma)$ as $\sigma \rightarrow 0$, resulting in $H(\vect{a}|\vect{s}, \vect{z}) \rightarrow - \infty$.
Therefore, we used TD3 as a based RL method.

\section{Related Work}
Learning from expert demonstrations is effective for obtaining complex and natural motions~\cite{Peng18}.
In terms of imitation learning, the problem of imitating diverse behaviors from expert demonstrations has been addressed in previous studies~\cite{Wang17,Li17,Sharma19,Merel19}.
In these methods, diverse behaviors are encoded in latent variables.
However, these imitation learning methods assume the availability of observations of diverse behaviors performed by experts.
In this study, instead of learning from demonstrations, we address the problem of encoding diverse behaviors by learning from a reward function through trial and error.

In the field of evolutionary computation, QD algorithms that identify diverse solutions have been investigated~\cite{Pugh16,Cully15,Parker-Holder20,Gangwani20}.
In these methods, multiple policies are trained to maximize both task-specific rewards and a behavior diversity metric.
For example, \cite{Gangwani20} demonstrated that multiple solutions can be found for robotic manipulation tasks using a QD algorithm.
However, these QD algorithms are typically limited to yielding a finite number of solutions. 
Our method can yield an infinite number of solutions by learning a continuous latent representation, thereby providing higher flexibility.
Learning such continuous latent representations allows us to continuously tune the behavior of the agent through low-dimensional representations.
Continuous control tasks often requires continuous tuning of the behavior of the agent, and continuous latent representations of diverse skills will be useful for downstream tasks.

Recently methods for learning diverse behaviors using deep RL with unsupervised rewards have been investigated~\cite{Eysenbach19,Sharma20}.
These methods can be used to learn a continuous skill space by training a latent-conditioned policy.
However, these methods are not aimed at learning a policy to solve a specific task, and they learn diverse behaviors randomly.
\cite{Kumar20} proposed a method Structured Maximum Entropy Reinforcement Learning (SMERL), which learns multiple solutions for a specified task.
However, information regarding the optimal return value before learning diverse solutions is required by that method. 
Therefore, a separate RL algorithm must be executed in advance to estimate the optimal return value because the optimal return value is typically unknown. 
By contrast, our method does not require information regarding the optimal return value, and a separate RL algorithm need not be executed.
In other studies~\cite{Burda19,Badia20, Badia20b}, the use of separate critics and different discount factors for extrinsic and intrinsic rewards in Atari games with discrete actions was investigated.
Our method can be regarded as an approach that leverages separate critics for extrinsic and intrinsic rewards with different discount factors.
Additionally, our method is applicable to problem domains with discrete actions by parameterizing $q(\vect{z}|\vect{s}, \vect{a})$ without an actor network.   

Hierarchical RL methods~\cite{Bacon17,Florensa17,Nachum18,Nachum19,Osa19} learn a hierarchical policy structured by a latent variable. 
However, the latent variable in HRL algorithms do not provide multiple solutions for a specified task.
Recently, \cite{Nachum18,Nachum19} investigated goal-conditioned policies $\pi(\vect{a}|\vect{s}, \vect{g}) $, where $\vect{g}$ denotes the goal. 
The goal-conditioned policy can be regarded as a method of modeling diverse behaviors conditioned on a latent variable that has semantic meaning. 
However, learning a goal-conditioned policy requires the user to specify the goals in a supervised manner. 
By contrast, we aim to learn diverse behaviors in an unsupervised manner by using only reward signals.

\begin{figure}[tb]
	\centering
	\subfigure[Hopper task.]{\includegraphics[width=0.19\columnwidth]{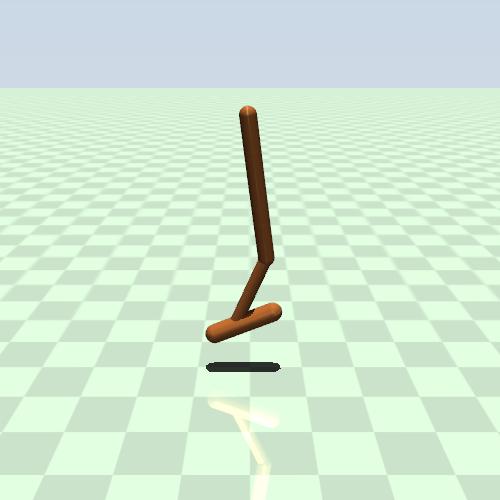}}
	\subfigure[Walker2d task.]{\includegraphics[width=0.19\columnwidth]{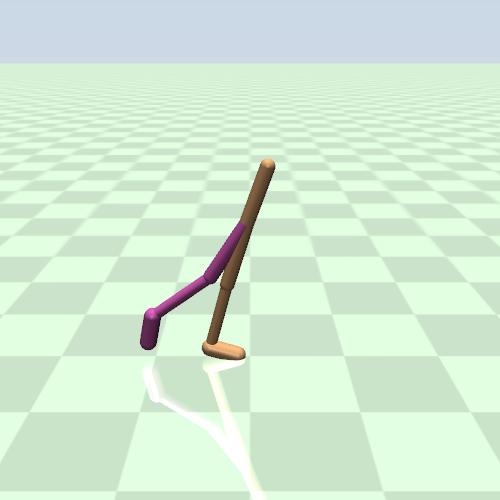}}
	\subfigure[Humanoid task.]{\includegraphics[width=0.19\columnwidth]{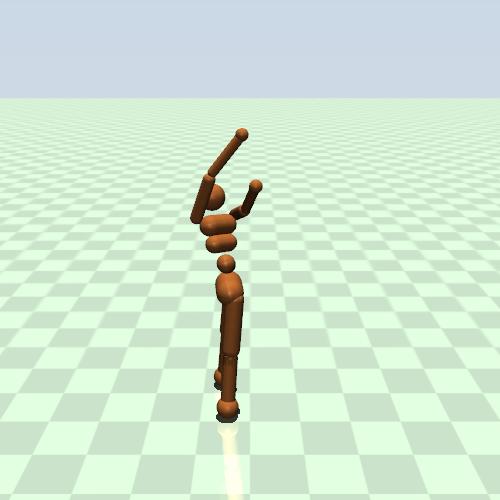}}
	\subfigure[Halfcheetah task.]{\includegraphics[width=0.19\columnwidth]{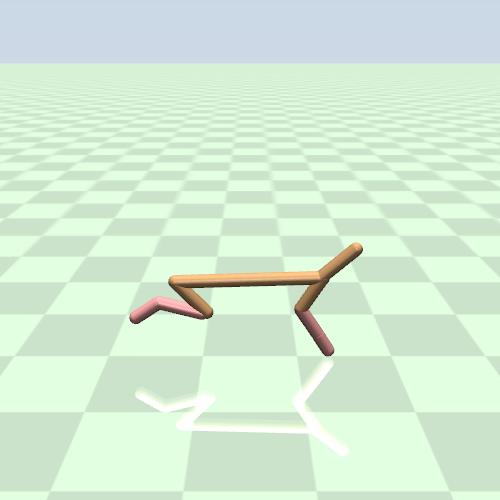}}
	\subfigure[Ant task.]{\includegraphics[width=0.19\columnwidth]{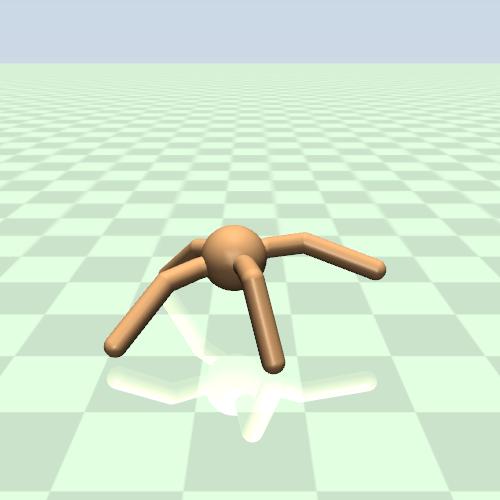}}
	\caption{Tasks used in experiments.}
	\label{fig:task}
	\vspace{1.0cm}
\end{figure}

\section{Experiments}
\label{sec:exp}
We performed experiments to evaluate 1)~the diversity of solutions obtained by LTD3, 2) effect of strategies for maximizing MI, and 3)~the few-shot robustness of policies obtained by LTD3.
We used tasks based on OpenAI Gym~\cite{brockman16} with MuJoCo physics engine~\cite{Todorov12}. 
The tasks were based on the Hopper, Walker2d, Humanoid, HalfCheetah, and Ant tasks,
as shown in Figure~\ref{fig:task}\footnote{Background color is changed from default setting for visibility}.

To emphasize the existence of diverse solution, a velocity term of the reward function is modified from that in the original tasks in OpenAI Gym.
In the reward function of the original tasks in OpenAI Gym, a velocity term $r_{\textrm{vel}}$ is used to encourage the agent to walk faster, and $r_{\textrm{vel}}$ is given by
\begin{align}
	r_{\textrm{vel}} = ( x_{t} - x_{t-1})) / \Delta t,
\end{align}
where $x_t$ represents the horizontal position of the agent at time $t$, and $\Delta t$ is the time step size in the simulation.
As in the study in \cite{Kumar20}, we modified the velocity term to the following: 
\begin{align}
	r_{\textrm{vel}} = \min\big(  (x_{t} - x_{t-1}) / \Delta t, v_{\textrm{max}} \big),
\end{align}
where $v_{\textrm{max}}$ is a constant term that defines the upper bound of $r_{\textrm{vel}}$.
Therefore, the agent is required to walk at the maximum speed $v_{\textrm{max}}$ with minimal control cost instead of walking as fast as possible.
We refer to the modified version of the Hopper, Walker2d and Humanoid tasks as HopperVel, Walker2dVel, and HumanoidVel tasks, respectively.
More details of these tasks are described in \ref{app:exp}.

We compared the proposed method with SMERL~\cite{Kumar20} and a variant of DIAYN~\cite{Eysenbach19}, which is denoted as SAC-DIAYN herein and is used as a baseline in the study in~\cite{Kumar20}. 
In SAC-DIAYN, the policy is trained to maximize the sum of the task reward and an unsupervised reward given by $r_{\textrm{diayn}} = \log q_{\vect{\psi}}(\vect{z}|\vect{s})$,
where $q_{\vect{\psi}}(\vect{z}|\vect{s})$ is a parameterized model that discriminates the value of the latent variable.
Although \cite{Kumar20} reported results based on discrete latent variables, this unsupervised reward can be applied to the continuous latent variable if a factored Gaussian distribution is assumed, as in \cite{Sharma20}.
In SMERL, $r_{\textrm{diayn}}$ is assigned only when the agent achieves a return that is close to the optimal return.
Therefore, an off-the-shelf RL method must be used to determine the optimal return before executing SMERL.
Our implementation of TD3 and SAC was adapted from the implementation in OpenAI Spinning Up~\cite{spinningup}.

We employed the diversity metric proposed in \cite{Parker-Holder20} to quantify the diversity of the learned solutions.
Given a set of policies $\{ \pi_i \}^M_{i=1}$, the diversity metric is given by
\begin{align}
	D_{\textrm{div}} = \det \left( K(\vect{\phi}(\pi_i), \vect{\phi}(\pi_j))^M_{i,j=1} \right),
\end{align}
where $\vect{\phi}(\pi) \in \R^{l}$ is the behavior embedding of policy $\pi$, and $K: \R^l \times \R^l \mapsto \R $ is a kernel function.
We used the squared-exponential kernel function given by 
\begin{align}
	k(\vect{\phi}_i, \vect{\phi}_j) = \exp \left( - \frac{ \left\| \vect{\phi}_i - \vect{\phi}_j \right\|^2 }{2h^2} \right),
\end{align}
where $h$ is the length scale, and $h > 0$. 
In our evaluation, we used $h=1000$ for the Ant task, and $h=100$ for the other tasks.
As described in \cite{Parker-Holder20}, we used the behavior embedding given by
\begin{align}
	\vect{\phi}(\pi) = \E_{\vect{s} \sim \pi, \mathcal{P}}\left[ \vect{a} | \pi(\vect{a}|\vect{s}), \vect{s}  \right],
\end{align}
which can be estimated using the states visited by policy $\pi$.
To evaluate the diversity of solutions obtained by learning the continuous latent variable, we sampled the value of $\vect{z}$ uniformly from $[-1, 1]$.

\subsection{Learning Diverse Solutions in Training MDPs}

%\begin{figure}[tb]
%	\subfigure[LTD3.]{\includegraphics[width=0.55\columnwidth]{walker2dvel_2d_seed5_comment}}
%	\subfigure[SAC-DIAYN.]{\includegraphics[width=0.45\columnwidth]{sac_diayn_walker2d_2d_seed3_intri_0.5}}
%	\caption{Diverse behaviors learned for Walker2dVel task. Latent variable is continuous and two-dimensional.}
%	\label{fig:Walker2dVel_ltd3}
%\end{figure}

\begin{figure}
	\centering
	\subfigure{\includegraphics[width=\columnwidth]{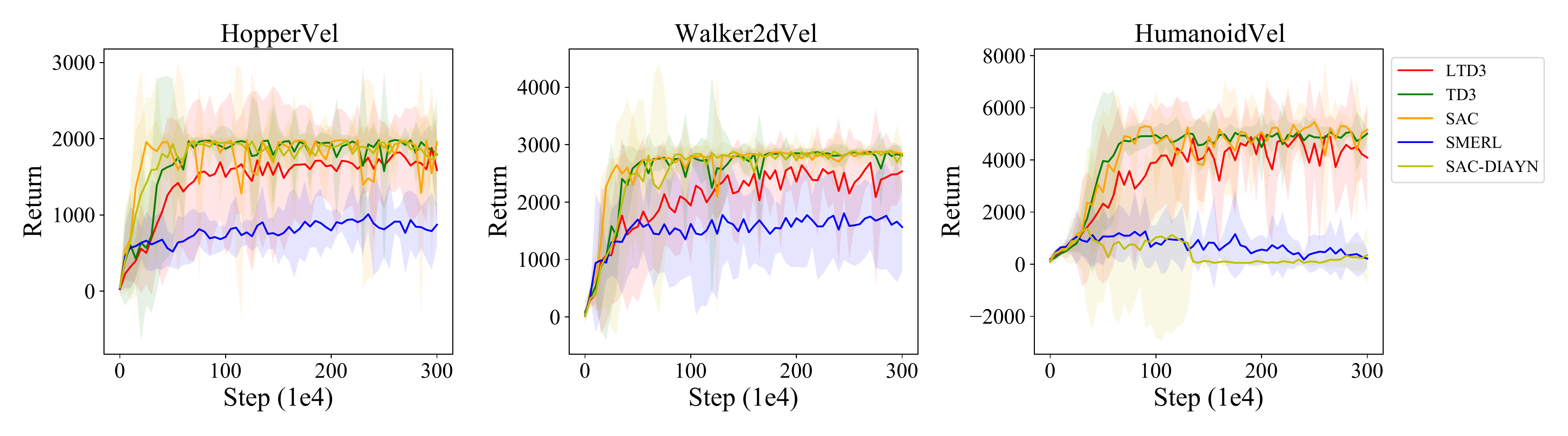}}
	\subfigure{\includegraphics[width=0.55\columnwidth]{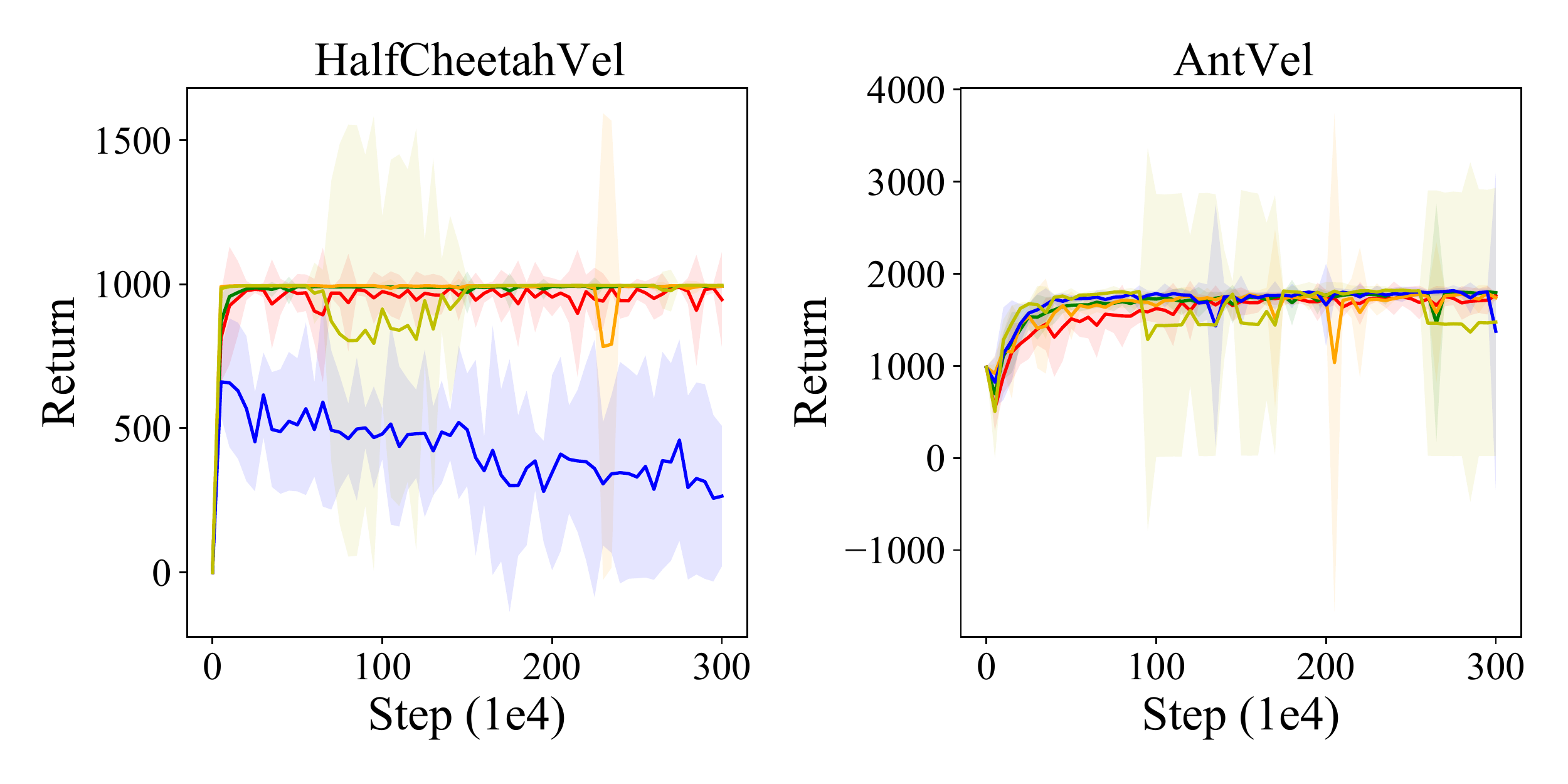}}
%	\vspace{-0.7cm}
	\caption{Comparison with baseline methods. In LTD3, SAC-DIAYN and SMERL, the latent variables are continuous and two-dimensional.}
	\label{fig:methods_learning_curve}
\end{figure}

\paragraph{Quantitative Comparison}
The learning curves of the baseline methods and LTD3 are shown in Figure~\ref{fig:methods_learning_curve}.
The results show that LTD3 is more sample-efficient than SMERL on these three tasks.
Compared with SAC-DIAYN, LTD3 is more sample-efficient than SAC-DIAYN on the HumanoidVel task.
Although SAC-DIAYN is slightly more sample-efficient than LTD3 on the Walker2dVel task, SAC-DIAYN did not obtain diverse solutions as discussed in the previous section.
The difference between LTD3 and other baseline methods for training a latent-conditioned policy is prominent in the HumanoidVel task, 
which indicates that LTD3 is advantageous over other methods when the state-action space is high dimensional.
These results show that LTD3 can obtain diverse solutions in a sample-efficient manner.
In \ref{app:ltd3}, we provide a detailed analysis of the effect of hyperparameters in LTD3.

The diversity scores are presented in Table~\ref{tbl:diversity_score}.
The diversity scores of the solutions obtained by LTD3 were higher than those obtained by SAC-DIAYN and SMERL. 
Although the only exception was the humanoid task, where the diversity score of SMERL was higher than that of LTD3, the return achieved by SMERL on the humanoid task was significantly lower than that achieved by LTD3. 
Moreover, the policy obtained by SMERL did not perform a meaningful walking behavior. 
These results indicate that LTD3 outperforms SMERL and SAC-DIAYN on these locomotion tasks in terms of both diversity and quality.

\begin{table}
	\caption{Diversity score of the solutions obtained by LTD3 and baseline methods.}
	\label{tbl:diversity_score}
	\centering
	\begin{tabular}{lccccc}
		\hline
		Task & Walker & Hopper & Humanoid & HalfCheetah & Ant \\
		\hline
		LTD3   & 0.79 $\pm$ 0.15  & 0.63 $\pm$ 0.33 & 0.20 $\pm$ 0.22 & 0.68 $\pm$ 0.26 & 0.46 $\pm$ 0.37 \\
		SMERL  & 0.66 $\pm$ 0.16 & 0.075 $\pm$ 0.092 & 0.44 $\pm$ 0.12 & $\ll$ 0.01 & $\ll$ 0.01 \\
		SAC-DIAYN & 0.01  $\pm$ 1.4 $\cdot 10^{-2}$  & $\ll$ 0.01  & $\ll$ 0.01   & $\ll$ 0.01& 0.06 $\pm$ 0.08  \\
		\hline
	\end{tabular}
	\vspace{0.2cm}
\end{table}

\begin{figure}[t]
%\begin{wrapfigure}{r}{0.6\textwidth}
	\subfigure[Results with two-dimensional continuous latent variable.]{\includegraphics[width=0.51\columnwidth]{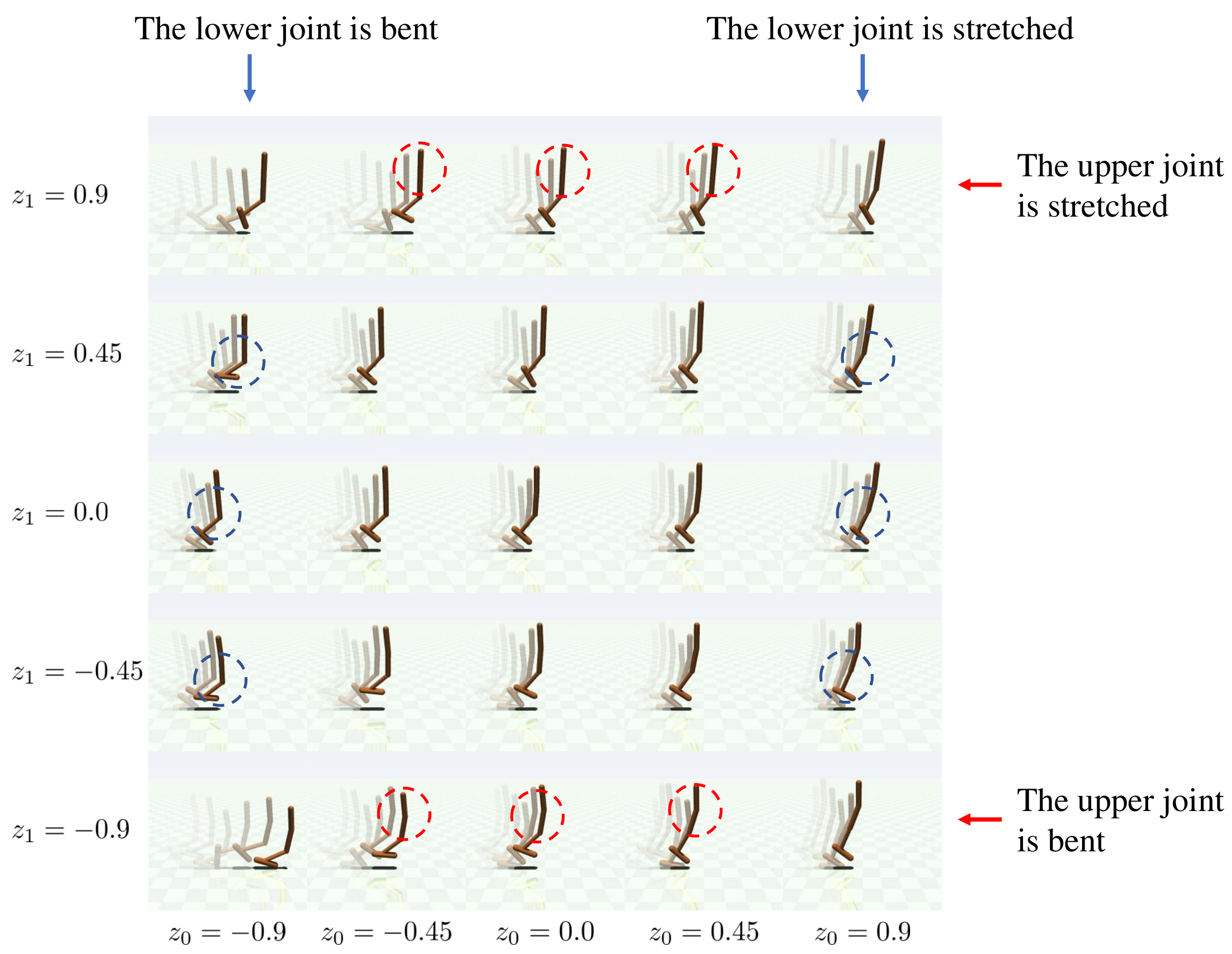}}
	\hfill
	\subfigure[Results with one-dimensional continuous latent variable and three categorical latent variable.]{\includegraphics[width=0.40\columnwidth]{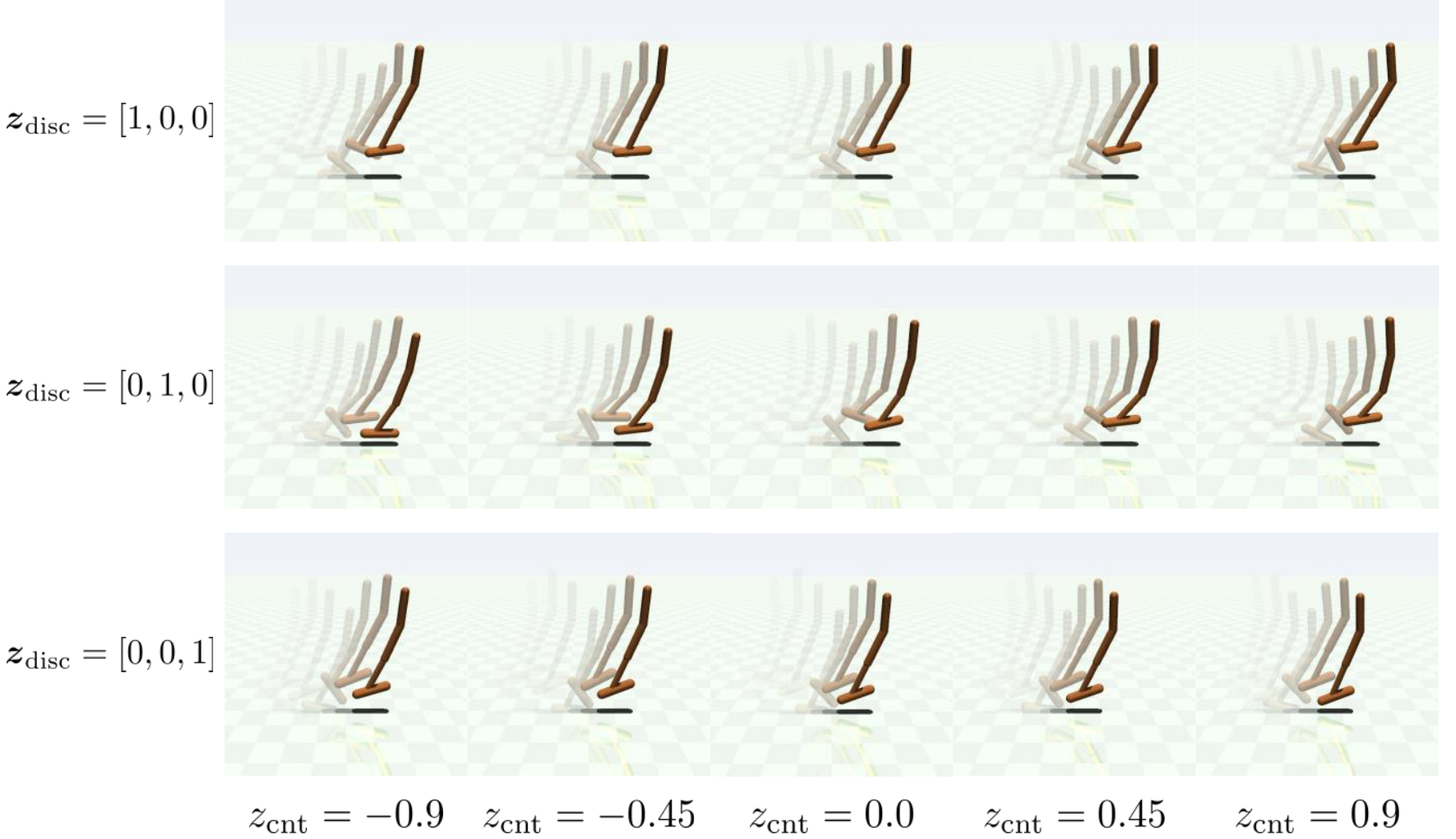}}
	\caption{Behavior learned by LTD3 for HopperVel task. }
	\label{fig:HopperVel}
	\end{figure}

\paragraph{Diverse Behaviors on HopperVel}
We show the behaviors learned by LTD3 on the HopperVel task in Figure~\ref{fig:HopperVel}. 
In these figure, the screens were capture at the same time steps, and the variance of the position indicated the variance of the walking speed.
As shown, LTD3 discovered diverse behaviors for solving given tasks.

%\end{wrapfigure}

Figure~\ref{fig:HopperVel}(a) shows the results with a two-dimensional continuous latent variable on the HopperVel task.
The agent on the HopperVel task has a leg with two knee joints, and the latent variable encoded various styles of using these joints for hopping.
The lower knee joint was bent when $z_0=-0.9$, whereas the lower knee joint was straight when $z_0 = 0.9$.
When $z_1=0.9$, the upper knee joint was straight, whereas the upper knee joint was bent if $z_1=-0.9$.
Figure~\ref{fig:HopperVel}(b) shows the results with a one-dimensional continuous latent variable and the three categorical latent variables on the HopperVel task.
When the discrete latent variable changed, the walking behavior changes discontinuously.
Regardless of the value of the discrete latent variable, the angle of the upper joint changed continuously when the value of the continuous latent variable changed continuously.
This result shows that LTD3 successfully learned diverse solutions that achieved the task, and that the user might select a preferred solution by selecting the value of the latent variable. 
It is challenging to manually design the reward function that specifically elicit one of these behaviors.

\begin{figure}[t]
	\centering
	\includegraphics[width=0.77\columnwidth]{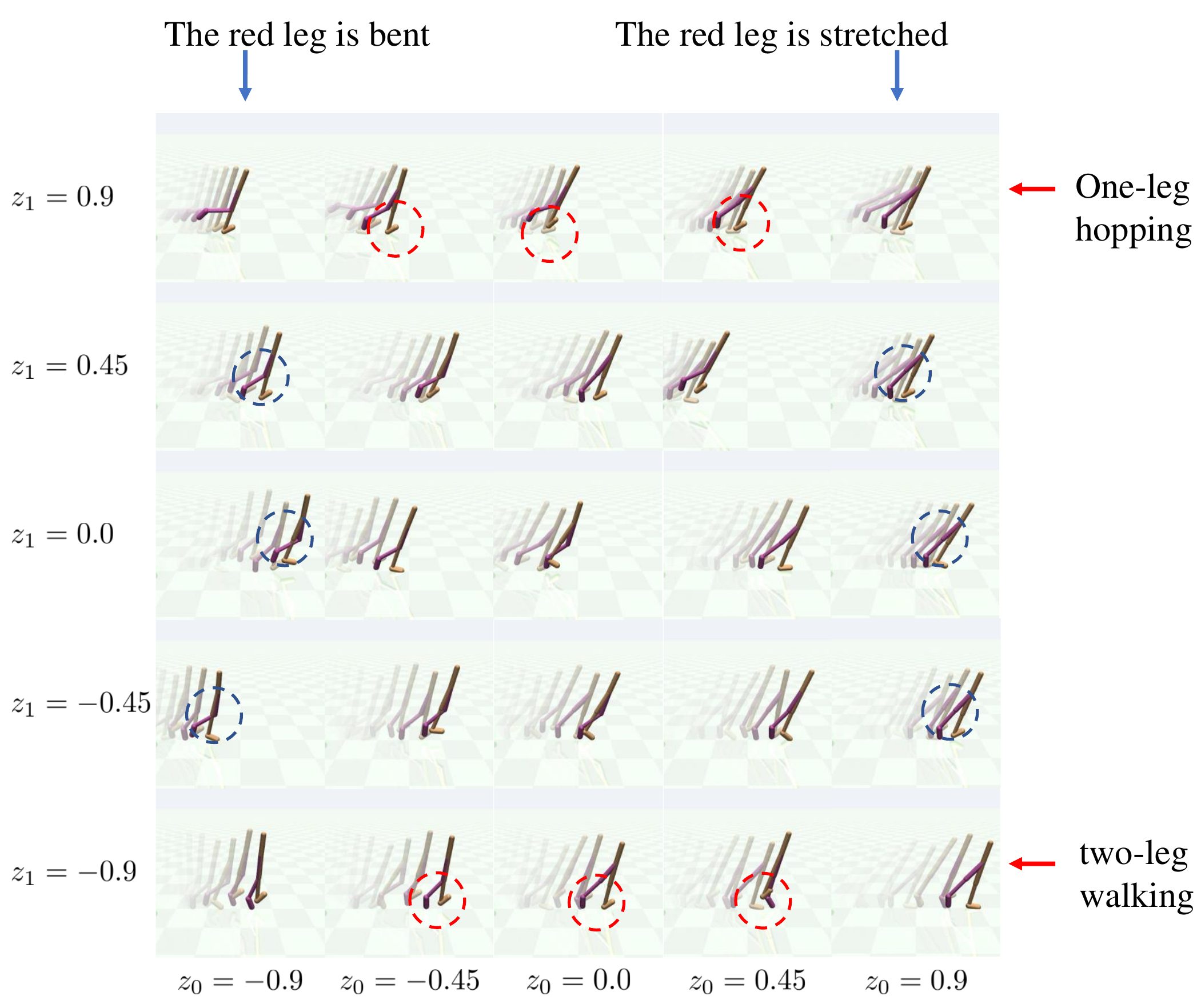}
	\caption{Diverse behaviors obtained by LTD3 for Walker2dVel task. Latent variable is continuous and two-dimensional.}
	\label{fig:Walker2dVel_ltd3}
\end{figure}

\begin{figure}[t]
	\centering
	\includegraphics[width=0.5\columnwidth]{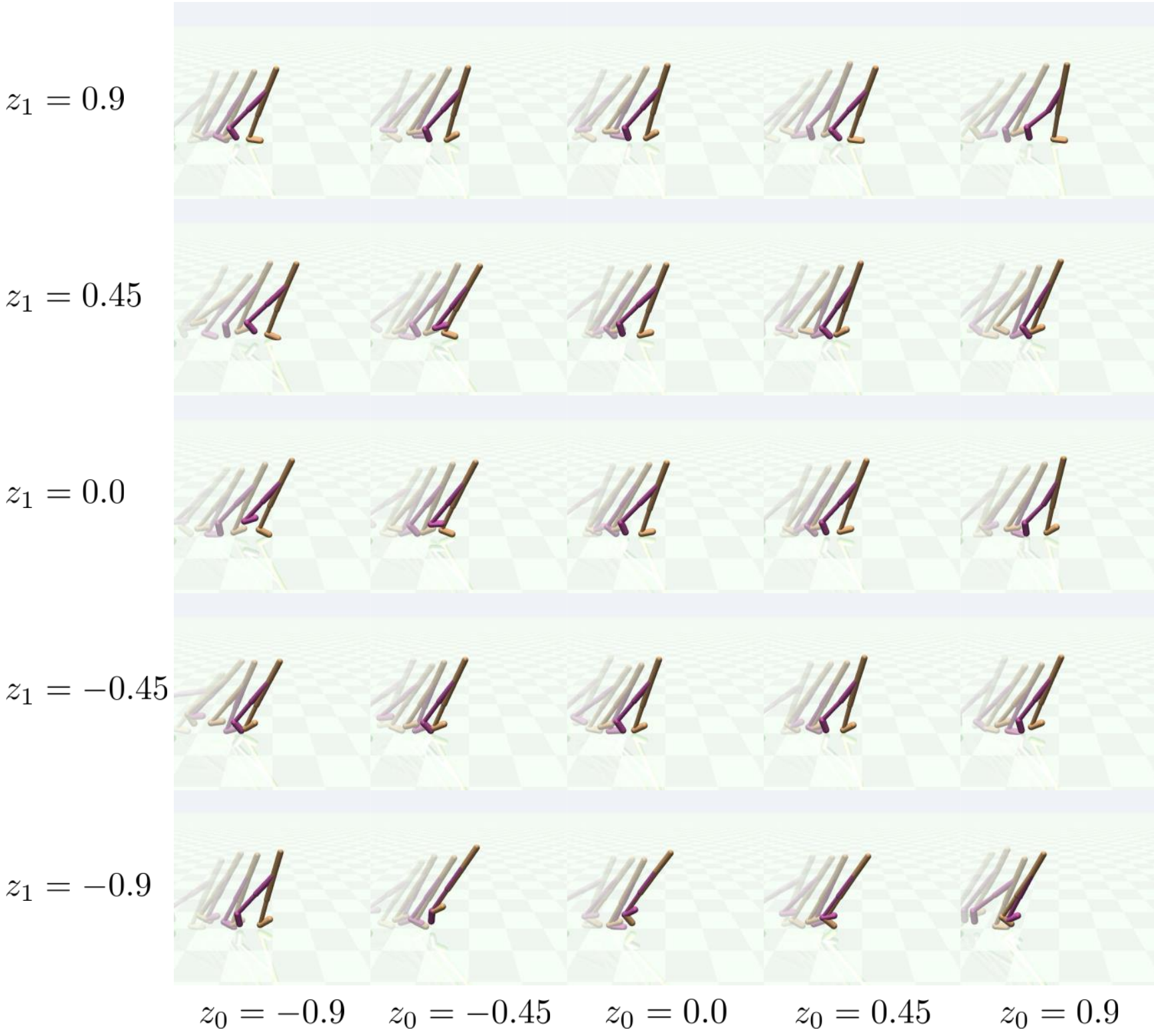}
	\caption{Diverse behaviors obtained by SAC-DIAYN for Walker2dVel task. Latent variable is continuous and two-dimensional.}
	\label{fig:Walker2dVel_sac_diayn}
\end{figure}

\paragraph{Qualitative comparison against a baseline method}
Figures~\ref{fig:Walker2dVel_ltd3} and~\ref{fig:Walker2dVel_sac_diayn} show the diverse walking behaviors obtained by LTD3 and SAC-DIAYN in the Walker2dVel task when the latent variable is continuous and two-dimensional.
In this figure, each frame shows an image of the walking behavior with different values of latent variables.
As shown in Figure~\ref{fig:Walker2dVel_ltd3}, the walking behavior obtained by LTD3 changes continuously according to the value of the latent variable.
The top row of Figure~\ref{fig:Walker2dVel_ltd3}, which corresponds to $z_1=0.9$, shows the agent performing one-leg hopping. 
Meanwhile, the bottom row of the figure, which corresponds to $z_1=-0.9$, shows the agent walking with two legs.
The left-most column of of Figure~\ref{fig:Walker2dVel_ltd3}, which corresponds to $z_0=-0.9$, shows the agent walking/hopping with the red leg bent. Meanwhile, the right-most column of the figure, which corresponds to $z_1=-0.9$, shows the agent walking/hopping with the red leg stretched.
By contrast, the policy obtained by SAC-DIAYN with a two-dimensional continuous latent variable did not exhibit such variation.

In Figure~\ref{fig:Walker2dVel_sac_diayn}, which shows the behaviors obtained by SAC-DIAYN for the Walker2dVel task, a clear difference in behaviors was not observed when the value of the latent variable was changed.
In SAC-DIAYN, we observed a tradeoff between the sample efficiency and solution diversity. Increasing the weight of the intrinsic reward enables more diverse behaviors to be achieved; however, the training becomes less sample efficient. 
Although we tuned the parameter to obtain diverse solutions with high sample efficiency in SAC-DIAYN, we did not obtain a parameter that can achieve the diversity and quality of solutions comparable to those obtained by LTD3.
The results indicates that LTD3 can obtain more diverse solutions than SAC-DIAYN.
In \ref{app:sac_diayn}, we provide a detailed analysis of the effect of weight on the intrinsic parameter in SAC-DIAYN.

\begin{figure}
	\centering
	\subfigure{\includegraphics[width=\columnwidth]{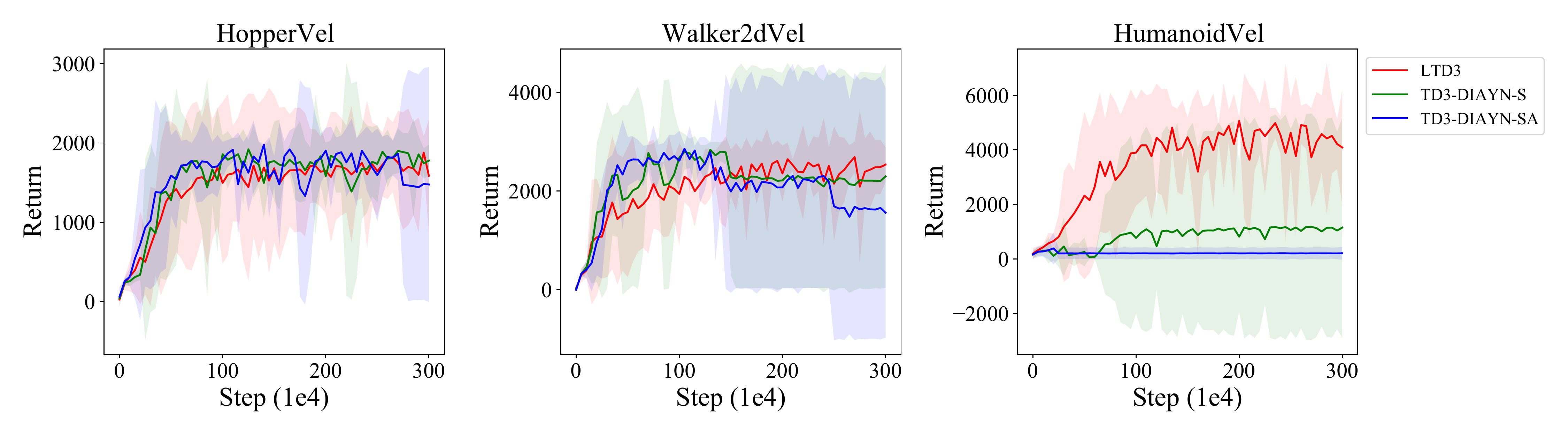}}
	\subfigure{\includegraphics[width=0.6\columnwidth]{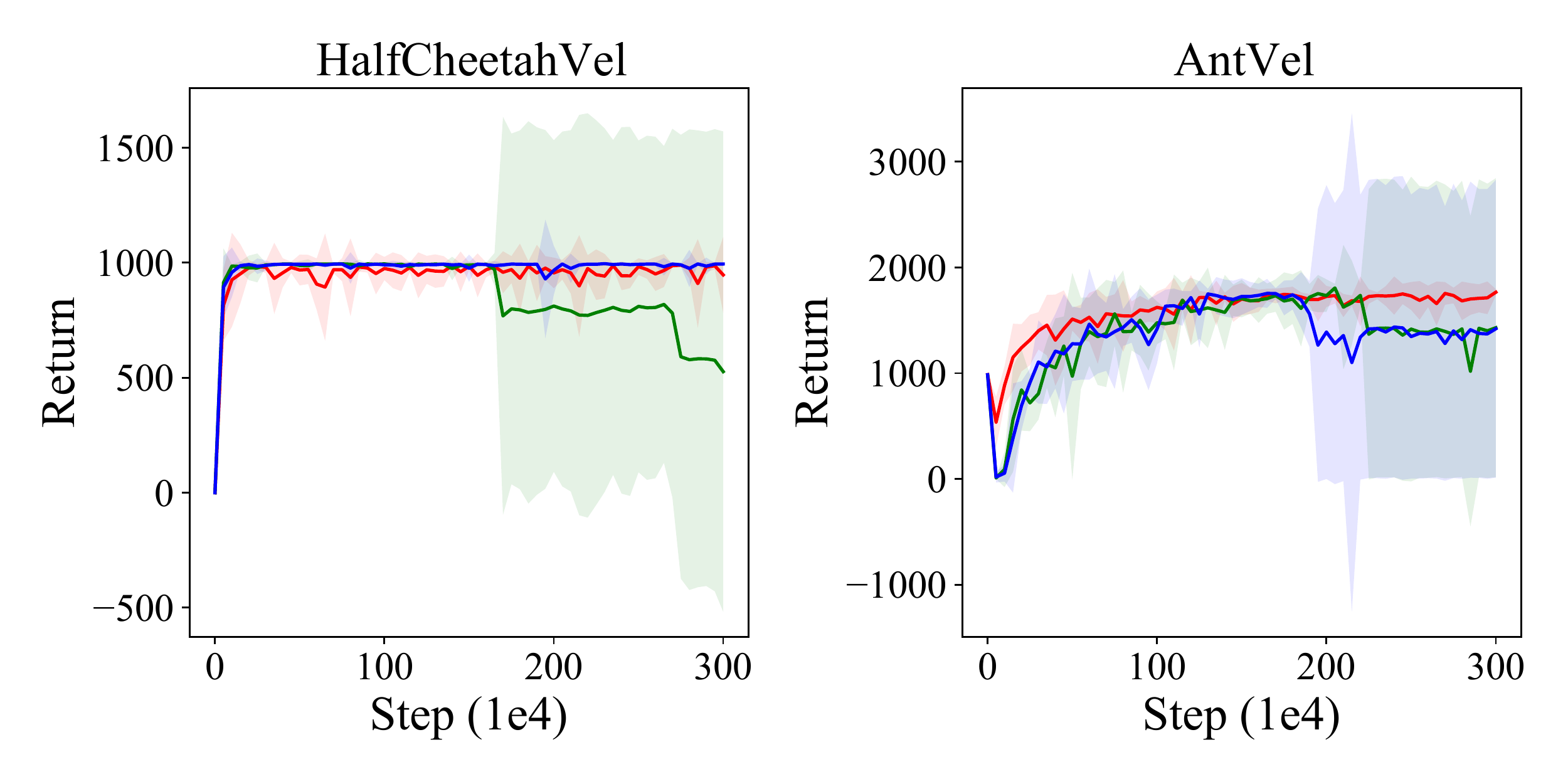}}
	%	\vspace{-0.7cm}
	\caption{Comparison of strategies for maximizing mutual information.}
	\label{fig:ltd3_ablation}
	%	\vspace{-0.3cm}
\end{figure}

\subsection{Comparison of strategies for maximizing MI}
Next, we analyze the improvement afforded by maximizing $I(\vect{s}, \vect{a} ; \vect{z})$ instead of $I(\vect{s} ; \vect{z})$. 
Specifically, we compare variants of TD3-based implementations of SAC-DIAYN, as shown in Figure~\ref{fig:ltd3_ablation}.
In a variant denoted as TD3-DIAYN-S, the unsupervised reward used in DIAYN $\log q_\vect{\phi}( \vect{z}| \vect{s})$ is maximized using TD3 in addition to the task reward. 
Similarly, in a variant denoted as TD3-DIAYN-SA, the unsupervised reward $r_{\textrm{diayn\_sa}} = \log q_\vect{\phi}( \vect{z}| \vect{s}, \vect{a})$, 
which is the variational lower bound of $I(\vect{s}, \vect{a} ; \vect{z})$, is maximized using TD3.

Note that the MI maximization through back-propagation cannot be used with a state-based MI. 
The variational lower bound of the state-based MI is given by $q(\vect{z}|\vect{s})$, which predicts the value of $\vect{z}$ for a given $\vect{s}$. 
Here, $q(\vect{z}|\vect{s})$ does not take the action a as the input. 
Therefore, the gradient from $q(\vect{z}|\vect{s})$ cannot be backpropagated to the actor network. 
The lower bound of the state-action-based MI is given by $q(\vect{z}|\vect{s}, \vect{a})$, not $q(\vect{z}|\vect{s})$. 
The approximated posterior distribution $q(\vect{z}|\vect{s}, \vect{a})$  takes the action generated by the actor as its input. 
Therefore, we can update the actor directly through backpropagation to maximize $q(\vect{z}|\vect{s}, \vect{a})$.

\begin{table}
	\caption{Diversity score of the solutions obtained by LTD3, TD3-DIAYN-S, and TD3-DIAYN-SA}
	\label{tbl:diversity_score_ablation}
	\centering
	\begin{tabular}{lccccc}
		\hline
		Task & Walker & Hopper & Humanoid & HalfCheetah & Ant \\
		\hline
		LTD3   & 0.79 $\pm$ 0.15  & 0.63 $\pm$ 0.33 & 0.20 $\pm$ 0.22 & 0.68 $\pm$ 0.26 & 0.46 $\pm$ 0.37 \\
		TD3-DIAYN-SA & 0.24  $\pm$ 0.25  & 0.21 $\pm$ 0.19  & $\ll$ 0.001 &  0.50 $\pm$ 0.27  & 0.37 $\pm$ 0.33\\
		TD3-DIAYN-S  & 0.35 $\pm$ 0.29 & 0.13 $\pm$ 0.14 & $\ll$ 0.001 & 0.40 $\pm$ 0.19  & 0.29 $\pm$ 0.19 \\
		\hline
	\end{tabular}
	%	\vspace{-0.2cm}
\end{table}

%We visualized the behaviors of the best policy among the five random seeds shown in Figure~\ref{fig:infomax_behavior}.
The learning curves are shown in Figure~\ref{fig:ltd3_ablation}, and the diversity scores of the obtained solutions are listed in Table~\ref{tbl:diversity_score_ablation}.
Among these variants, LTD3 exhibited the most diverse behaviors in the most stable manner.
The difference in the performance between LTD3 and TD3-DIAYN-SA resulted from the difference in the strategy for maximizing $\log q_\vect{\phi}( \vect{z}| \vect{s}, \vect{a})$.
When we deal with $\log q_\vect{\phi}( \vect{z}| \vect{s}, \vect{a})$ as an unsupervised reward, the performance is affected by the error in approximating the value function~\cite{Fujimoto18}.
The gradient estimator using function approximation is biased unless compatible function approximation is used~\cite{Sutton99}.  
By contrast, in LTD3, we can directly compute the unbiased gradient of the objective function which represents the tight lower bound of the MI.

Regarding the learning curve shown in Figure~\ref{fig:ltd3_ablation}, 
although the means of the returns achieved by TD3-DIAYN-S and TD3-DIAYN-SA were comparable to those of LTD3 on HopperVel and WalkerVel tasks, the variance was significantly high; this indicates that the learning with TD3-DIAYN-S and TD3-DIAYN-SA was unstable compared with LTD3. 
The advantage of LTD3 over TD3-DIAYN-S and TD3-DIAYN-SA was evident in the HumanoidVel task, where the state space is high dimensional.
While the sample efficiencies of TD3-DIAYN-S and TD3-DIAYN-SA are comparable, the diversity score of TD3-DIAYN-SA is higher than that of TD3-DIAYN-S on three tasks out of five. 
However, TD3-DIAYN-S outperforms TD3-DIAYN-SA on the walker task, which indicates that the comparison of $I(\vect{s},\vect{a};\vect{z})$ and $I(\vect{s};\vect{z})$ is dependent on the task properties. 
In terms of the diversity score and the achieved returns, LTD3 outperforms TD3-DIAYN-S and TD3-DIAYN-SA. 
The results indicate that both 1) state-action-MI and 2) MI maximization through backpropagation are necessary to achieve a satisfactory performance.

\subsection{Few-shot Robustness}
To further demonstrate the usefulness of our method, we evaluate the few-shot robustness of the policy trained using LTD3, based on the protocol proposed in \cite{Kumar20}.
A policy was trained on a training environment for three million time steps, and the trained policy was adapted to the test environment in 25 episodes as shown in Figure~\ref{fig:intro}.
In this experiment, we evaluated LTD3, SAC-DIAYN and SMERL with two-dimensional continuous, and discrete latent variables.
More details pertaining to the few-shot adaptation experiment are provided in \ref{app:fewshot}.

The results of the few-shot adaptation tasks with the Walker2d agent are presented in Table~\ref{tbl:few-shot}.
The high variance of returns in test MDPs was due to the property of the few-shot adaptation task, i.e., 
all methods can achieve high returns because a skill suitable for test MDPs may be obtained in a training MDP by chance. 
Only when the skills suitable for test MDPs are obtained with all random seeds can the mean of the returns be high with a low variance.
Table~\ref{tbl:few-shot} shows that LTD3 outperformed SMERL and SAC-DIAYN with statistically significant differences.
The results of the few-shot adaptation experiment indicate that LTD3 yields more diverse solutions than SMERL and SAC-DIAYN for both continuous and discrete latent variables.
In addition, SMERL required executing an off-the-shelf RL method to approximate the return of the optimal policy, which is not required in LTD3.
Therefore, LTD3 is more sample-efficient than SMERL in the training phase.

\begin{figure}
	\centering
	\subfigure[W-Short1.]{\includegraphics[width=0.2\columnwidth]{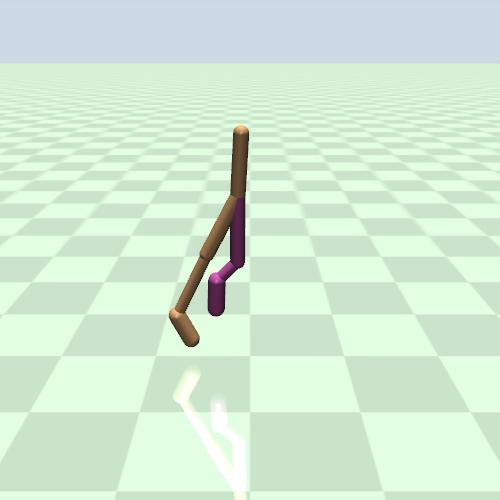}}
	\subfigure[W-Short2.]{\includegraphics[width=0.2\columnwidth]{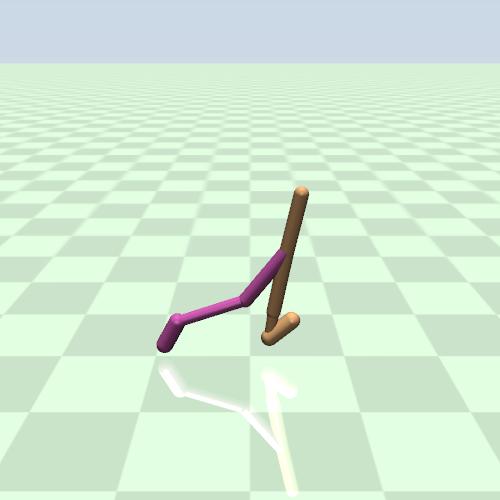}}
	\subfigure[W-LowShort.]{\includegraphics[width=0.2\columnwidth]{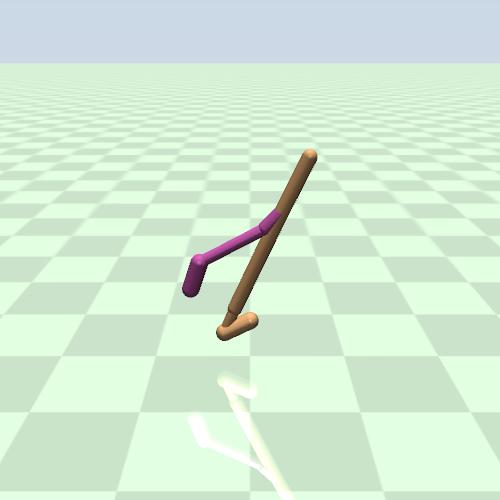}}
	\subfigure[W-ShortHigh.]{\includegraphics[width=0.2\columnwidth]{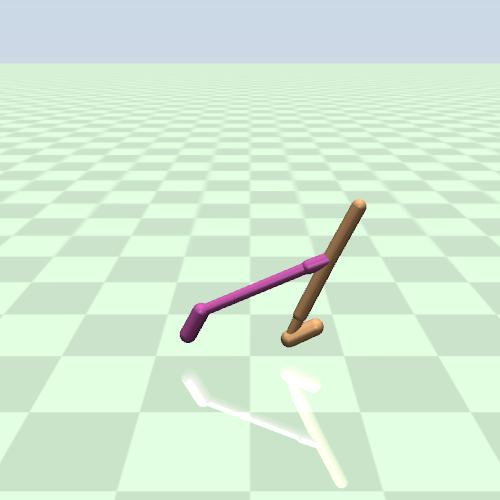}}
%	\vspace{-0.5cm}
	\caption{Tasks used in few-shot adaptation experiments.}
%	\vspace{-0.5cm}
	\label{fig:Few-shot_task}
\end{figure}

\begin{table}
	\caption{Results of few-shot adaptation tasks. The mean and standard deviation of the return in test MDPs are shown.}
	\label{tbl:few-shot}
	\centering
	\begin{tabular}{lcccc}
		\hline
		Test MDP & W-Short1 & W-Short2 & W-LowShort & W-ShortHigh  \\
		\hline
		LTD3 (disc) & \textbf{1519.8 $\pm$ 948.1} & \textbf{1685.0 $\pm$ 963.8} &\textbf{ 804.2 $\pm$ 953.2} & \textbf{1090.8 $\pm$ 783.3}  \\
		SMERL (Disc) & 803.3 $\pm$ 197.5 & 1050.5 $\pm$ 664.6 & 509.8 $\pm$ 354.7 & 873.5 $\pm$ 453.2 \\
		SAC-DIAYN (disc) & 621.7 $\pm$ 420.8 & 662.6 $\pm$ 408.1 & 390.4 $\pm$ 321.1 & 726.2 $\pm$ 637.6 \\ 
		LTD3 (2d)   & \textbf{1360.9 $\pm$ 1040.8} &\textbf{ 1585.0 $\pm$ 876.4} & 361.4 $\pm$ 111.7 & 800.2 $\pm$ 565.6  \\
		SMERL (2d) & 663.7 $\pm$ 304.9 & 656.9 $\pm$ 504.2 & 389. $\pm$  236.8 & 721.2 $\pm$ 423.9 \\
		SAC-DIAYN (2d) & 435.7 $\pm$ 224.7 &  890.5 $\pm$ 824.2 & 313.5 $\pm$ 155.8 &  759.9 $\pm$ 596.7\\
		\hline
	\end{tabular}
%	\vspace{-0.2cm}
\end{table}

\section{Discussion}
Although we demonstrated the advantage of the proposed method experimentally, the tasks used in our experiments are related to ``how to walk'' rather than ``where to walk,'' in which one can expect that actions contain meaningful information to obtain diverse behaviors. 
As demonstrated in previous studies~\cite{Eysenbach19,Kumar20,Sharma20}, when state $\vect{s}$ contains sufficient information to encourage necessary diversity, maximizing $I(\vect{s};\vect{z})$ is suitable.

In many off-policy actor-critic methods such as TD3 and SAC, the gradient with respect to the policy parameter is computed to maximize $\E[Q_\vect{w} (\vect{s},\vect{f}_\vect{\theta} (\vect{s}))]$, where $Q_\vect{w} (\vect{s}, \vect{a})$ is an approximated Q-function and $\vect{f}_\vect{\theta} (\vect{s})$ is a function that represents a policy and generates actions. If there is an approximation error, the estimated gradient is biased unless the compatible function approximation is used~\cite{Sutton99,Silver14}. Furthermore, the error of the function approximation significantly affects the learning performance of actor-critic methods, as indicated in \cite{Fujimoto18}.  

Regarding MI as an unsupervised reward, it is maximized based on the approximation of the value function. 
In the methods adopted in~\cite{Kumar20, Sharma20}, the unsupervised reward term $r_\textrm{info}$ is maximized using the critic as the task-specific reward term $ r_\textrm{task}$. 
As the reward function is given by $ r= r_\textrm{task}+ r_\textrm{info}$, the objective function for optimizing the policy in SAC-DIAYN can be decomposed as
\begin{align}
	\E\left[ Q_\vect{w} (\vect{s},\vect{f}_\vect{\theta} (\vect{s})) \right] = \E\left[ Q^{\textrm{task}}_\vect{w} (\vect{s},\vect{f}_\vect{\theta} (\vect{s})) \right] + \E\left[ Q^{\textrm{info}}_\vect{w} (\vect{s},\vect{f}_\vect{\theta} (\vect{s})) \right],
	\label{eq:sac_diayn_obj}
\end{align}
where $\E\left[ Q^{\textrm{task}}_\vect{w} (\vect{s},\vect{f}_\vect{\theta} (\vect{s})) \right]$ is the expected return that corresponds to the task-specific reward $r_\textrm{task}$, and $\E\left[ Q^{\textrm{info}}_\vect{w} (\vect{s},\vect{f}_\vect{\theta} (\vect{s})) \right]$ is the expected return that corresponds to the unsupervised reward $r_\textrm{info}$.
The objective function described in \eqref{eq:sac_diayn_obj} for SAC-DIAYN corresponds to \eqref{eq:objective} in the proposed method. 
As the maximization of $\E\left[ Q^{\textrm{info}}_\vect{w} (\vect{s},\vect{f}_\vect{\theta} (\vect{s})) \right]$ involves the appropriate Q-function, the estimated gradient with respect to the policy parameter is biased, as in many deep actor-critic methods.
In our method, instead of $\E\left[ Q^{\textrm{info}}_\vect{w} (\vect{s},\vect{f}_\vect{\theta} (\vect{s})) \right]$, we maximize $\E\left[ q_\vect{\phi} (\vect{z} | \vect{s},\vect{f}_\vect{\theta} (\vect{s})) \right]$, which is equivalent to performing a maximum likelihood estimation (MLE) of $q(\vect{z} | \vect{s}, \vect{a})$ that predicts the value of the latent variable z for given s and a. 
MLE is an unbiased estimator; therefore, we can update the policy parameter to maximize the variational lower bound of the mutual information using the unbiased gradient.

In addition, DIAYN and SMERL maximize the cumulative sum of the unsupervised reward approximated by the Q-function based on the SAC, not the immediate reward. 
However, it is not clear whether maximizing the cumulative unsupervised reward is necessary because it is not clearly justified in previous studies ~\cite{Eysenbach19,Kumar20,Sharma20}. 
The unsupervised reward is derived as the lower bound of the mutual information~\cite{Eysenbach19}, but the lower bound of the mutual information is given by the expected value of $q(\vect{z}|\vect{s})$, not the expected value of the discounted sum of $q(\vect{z}|\vect{s})$ over future time steps. 
However, DIAYN and SMERL actually maximize the discounted sum of $q(\vect{z}|\vect{s})$ using SAC. 
Although maximizing the discounted sum of $q(\vect{z}|\vect{s})$ may lead to maximizing $q(\vect{z}|\vect{s})$,there may be a slight difference between the bound of the mutual information $I(\vect{z}|\vect{s})$ and what DIAYN and SMERL actually optimize.

As we discussed, the discounted sum of $q(\vect{z}|\vect{s},\vect{a})$ does not appear in the objective function, and the derivation indicates that we need to maximize the expected value of $q(\vect{z}|\vect{s},\vect{a})$. 
Based on these observations, we proposed to maximize $\E\left[ q_\vect{\phi} (\vect{z} | \vect{s},\vect{f}_\vect{\theta} (\vect{s})) \right]$, which enables us to directly update the actor through backpropagation. The experimental results demonstrate that the proposed algorithm, LTD3 outperforms baseline methods in locomotion tasks, which we believe supports the abovementioned discussion.

\section{Conclusions}
We presented LTD3, which can train a latent-conditioned policy to represent diverse solutions.
In our approach, diverse behaviors are encoded in continuous or discrete latent variables via variational information maximization.
To maximize MI in a sample-efficient manner, the truncated importance sampling technique was proposed.
The experimental results indicated that our method can learn diverse solutions in continuous control tasks by learning a continuous skill space.
The effectiveness of our method in terms of few-shot robustness was empirically evaluated, and our method outperformed baseline methods in these settings. 
In future work, we will investigate methods to leverage diverse solutions in various applications.

\section*{Acknowledgement}
TO was supported by KAKENHI 19K20370, and MS was supported by KAKENHI 17H00757.

\appendix
\section{Hyperparameter of SAC-DIAYN}
\label{app:sac_diayn}
\begin{figure}[b]
	\subfigure{\includegraphics[width=\columnwidth]{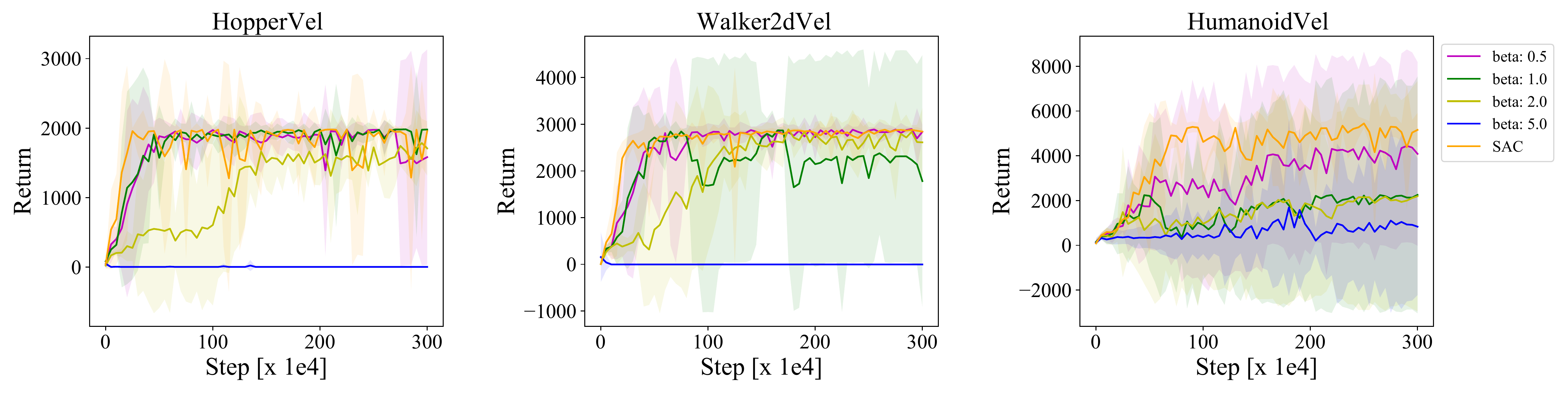}}
	\vspace{-0.7cm}
	\caption{Effect of coefficient for intrinsic reward in SAC-DIAYN. Latent variables are continuous and two-dimensional.}
	\label{fig:sac_diayn_intricoeff}
\end{figure}

\begin{figure}
	\subfigure[$\beta=0.5$.]{\includegraphics[width=0.32\columnwidth]{sac_diayn_walker2d_2d_seed3_intri_0.5}}
	\subfigure[$\beta=1.0$.]{\includegraphics[width=0.32\columnwidth]{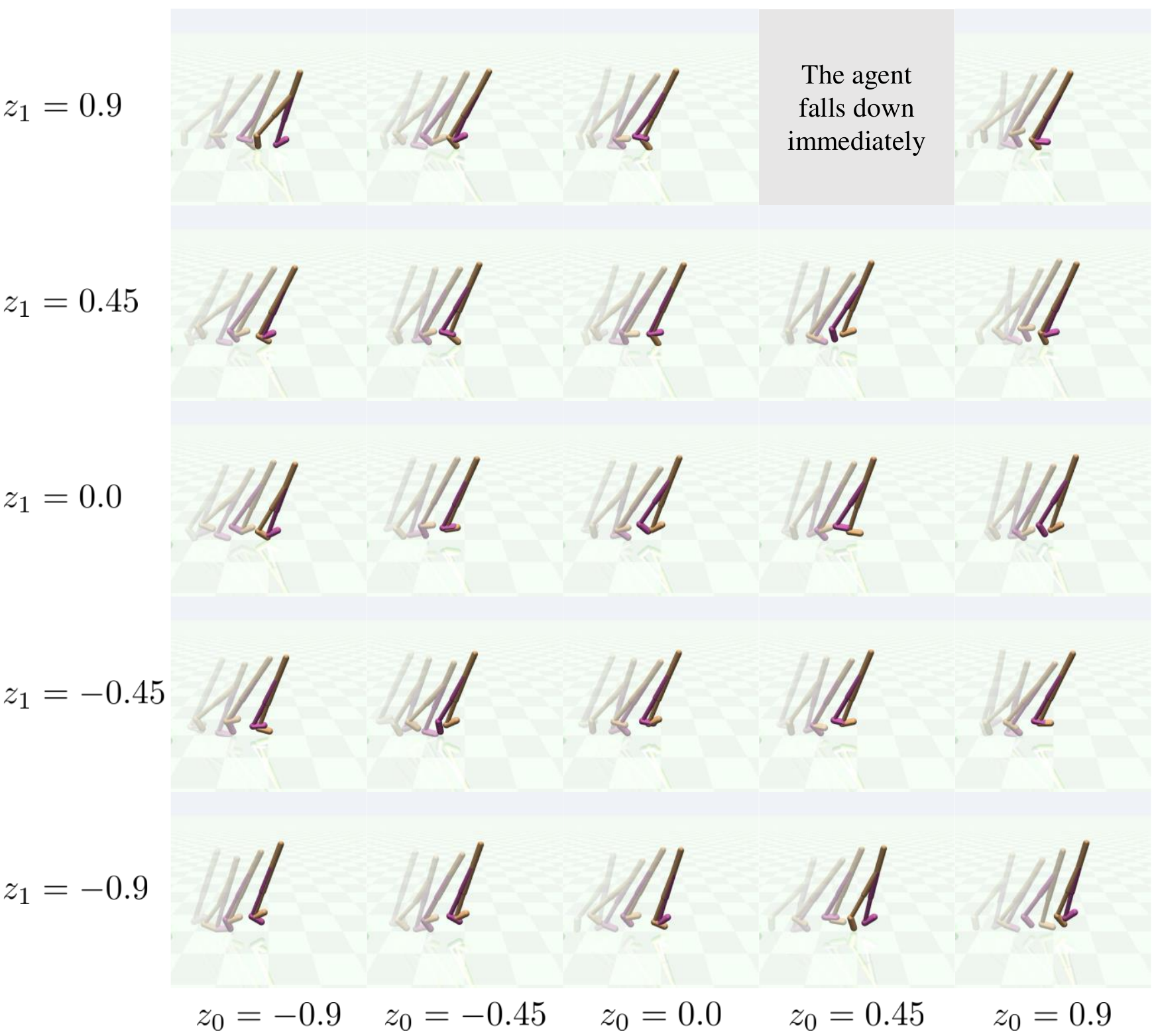}}
	\subfigure[$\beta=2.0$.]{\includegraphics[width=0.32\columnwidth]{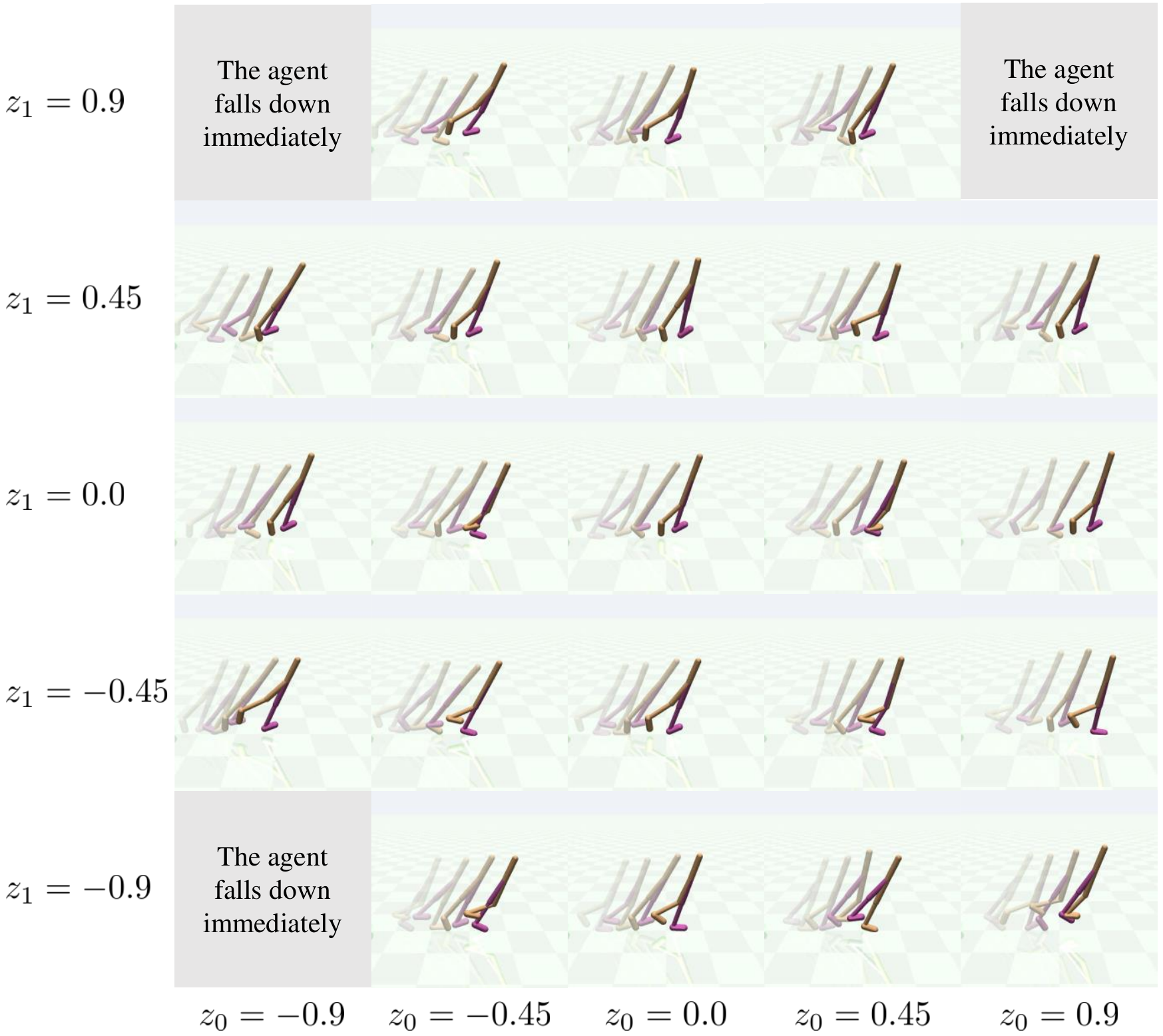}}
	\caption{Effect of coefficient for intrinsic reward in SAC-DIAYN. Latent variables are continuous and two-dimensional.}
	\label{fig:sac_diayn_intricoeff_behavior}
\end{figure}

A trade-off exists between the diversity of solutions and sample-efficiency in SAC-DIAYN.
In SAC-DIAYN, we maximize the reward
\begin{align}
	r = r_{\textrm{task}} + \beta r_{\textrm{diayn}},
\end{align}
where $r_{\textrm{task}}$ is a term encourage the behavior specific to a given task, $r_{\textrm{diayn}} = \log q_{\vect{\psi}}(\vect{z}|\vect{s})$ is the unsupervised reward, 
and $\beta$ controls the balance between $r_{\textrm{task}}$ and $r_{\textrm{diayn}}$.
When $\beta=0.5$ and $\beta=1.0$, the sample-efficiency of SAC-DIAYN is comparable to that of TD3 on the HopperVel and Walker2dVel tasks, as shown in Figure~\ref{fig:sac_diayn_intricoeff}.
However, the behavior of the agent does not change significantly even when the value of the latent variable is changed, as shown in Figures~\ref{fig:sac_diayn_intricoeff_behavior}(a) and (b).
When $\beta=2.0$, some variations are observed in the walking behaviors as Figure~\ref{fig:sac_diayn_intricoeff_behavior}(c); however, the training is less sample-efficient compared with when $\beta=0.5$ and $\beta=1.0$.
When $\beta=5.0$, we could not train a policy to achieve high returns in the specified tasks, as shown in Figure~\ref{fig:sac_diayn_intricoeff}.

Next, we provide the results of the few-shot adaptation tasks of SAC-DIAYN for different values of $\beta$.
Although the visualization of the obtained behavior indicates that more diverse behaviors were learned with $\beta=2.0$ than with $\beta=0.5$, SAC-DIAYN showed the best performance when $\beta=0.5$.
This results indicate that both robustness and diversity are necessary for achieving high performance in the few-shot adaptation task.
Although we tuned the parameter $\beta$ for SAC-DIAYN as described above, LTD3 still outperformed SAC-DIAYN when $\beta=0.5$.

\begin{table}
	\caption{Results of few-shot adaptation tasks of SAC-DIAYN. The mean and standard deviation of the return in test MDPs are shown.}
	\label{tbl:few-shot-sac-diayn}
	\centering
	\begin{tabular}{lcccc}
		\hline
		Test MDP & W-Short1 & W-Short2 & W-LowShort & W-ShortHigh  \\
		\hline
		$\beta=0.5$ (2d) & 435.7 $\pm$ 224.7 &  890.5 $\pm$ 824.2 & 313.5 $\pm$ 155.8 &  759.9 $\pm$ 596.7\\
		$\beta=0.5$ (disc) & 621.7 $\pm$ 420.8 & 662.6 $\pm$ 408.1 & 390.4 $\pm$ 321.1 & 726.2 $\pm$ 637.6 \\
		$\beta=1.0$ (2d) & 266.1 $\pm$ 212.7 & 990.5 $\pm$ 1080.5 & 182.8 $\pm$ 147.8 & 854.9 $\pm$ 1031.6 \\
		$\beta=1.0$ (disc) & -0.8 $\pm$  0.1 & -0.2 $\pm$ 0.1 & 4.1 $\pm$ 12.4 &  0.0 $\pm$  0.1  \\ 
		$\beta=2.0$ (2d) & 476.0 $\pm$  273.8& 440.9 $\pm$ 228.7 & 335.5 $\pm$ 189.1 & 351.9 $\pm$ 236.0 \\
		$\beta=2.0$ (disc) & -0.7 $\pm$ 0.3 & -0.1 $\pm$ 0.2 & -0.7 $\pm$ 0.2 & -0.1 $\pm$ 0.1  \\ 
		\hline
		LTD3 (2d)   & \textbf{1360.9 $\pm$ 1040.8} &\textbf{ 1585.0 $\pm$ 876.4} & 361.4 $\pm$ 111.7 & 800.2 $\pm$ 565.6  \\
		LTD3 (disc) & \textbf{1519.8 $\pm$ 948.1} & \textbf{1685.0 $\pm$ 963.8} &\textbf{ 804.2 $\pm$ 953.2} & \textbf{1090.8 $\pm$ 783.3}  \\
		\hline
	\end{tabular}
\end{table}

\begin{table}
	\caption{Diversity score of the solutions obtained by SAC-DIAYN for the Walker2dVel task with different values of $\beta$.}
	\label{tbl:diversity_score_diayn}
	\centering
	\begin{tabular}{lc}
		\hline
		& Diversity Score\\
		\hline
		 $\beta = 0.5$ & 9.9 $\cdot 10^{-3}$  $\pm$ 1.4 $\cdot 10^{-2}$\\
		 $\beta = 1.0$ & 0.035  $\pm$ 0.049 \\
		 $\beta = 2.0$ & 0.090  $\pm$ 0.072 \\
		\hline
	\end{tabular}
	%	\vspace{-0.2cm}
\end{table}

\section{Hyperparameter of LTD3}
\label{app:ltd3}

\subsection{Clipping Parameter for Importance Weight}
\begin{figure}
	\centering
	\subfigure{\includegraphics[width=0.32\columnwidth]{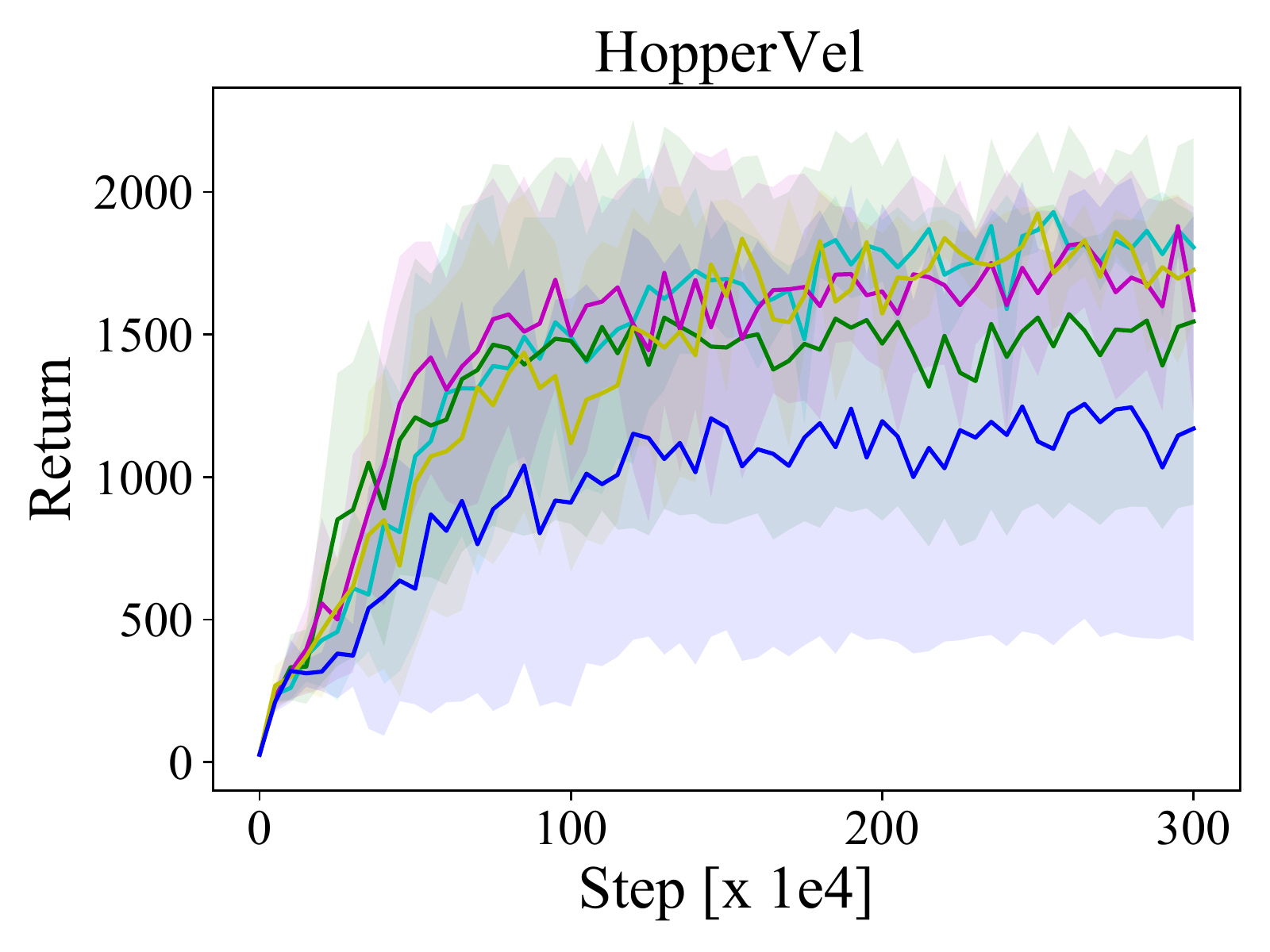}}
	\subfigure{\includegraphics[width=0.32\columnwidth]{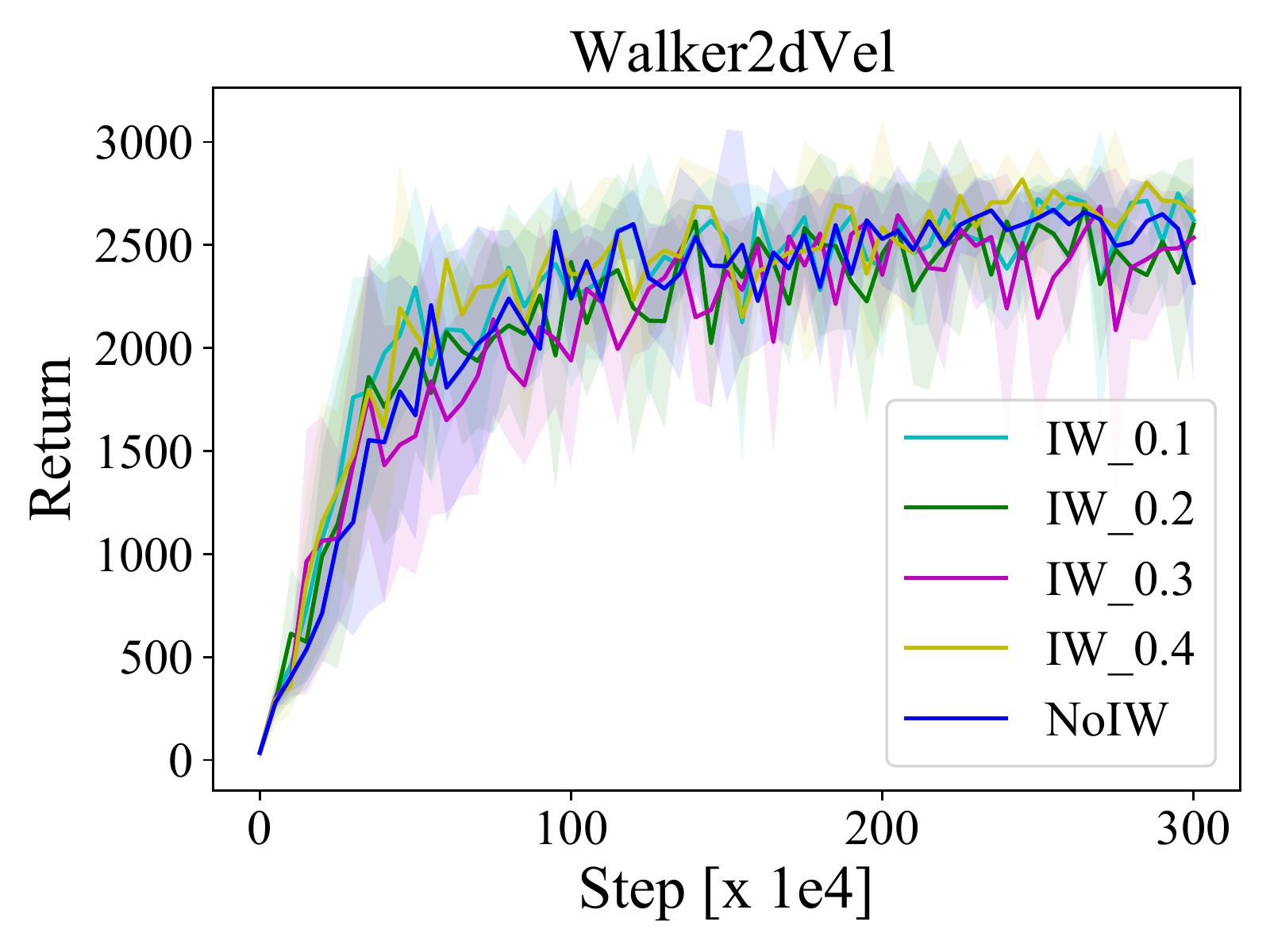}}
	\subfigure{\includegraphics[width=0.32\columnwidth]{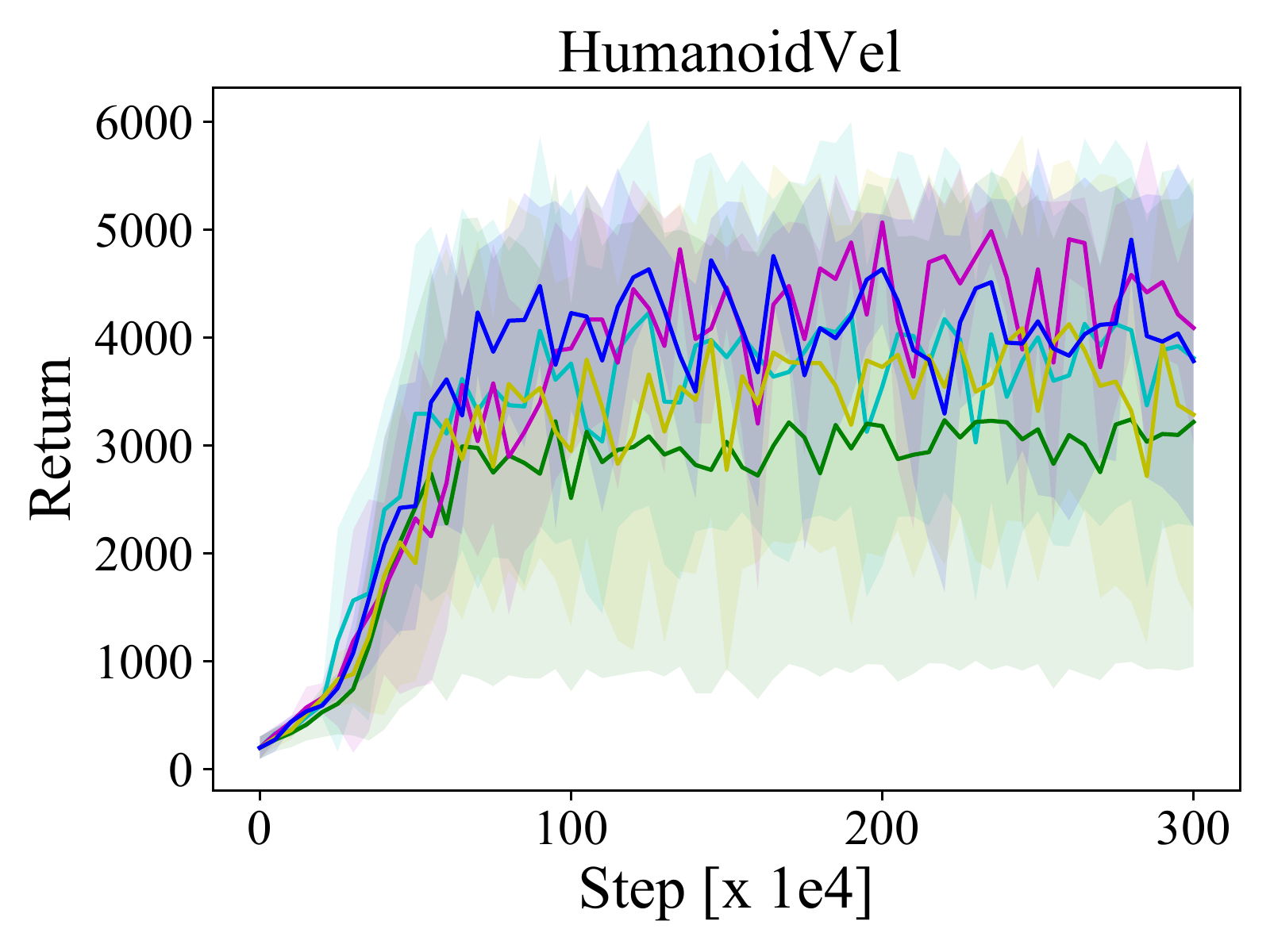}}
	\caption{Effect of clipping parameter on LTD3.
		Results the two-dimensional continuous latent variable.}
	\label{fig:IW_app}
\end{figure}

The learning curves of LTD3 with different values of the clipping parameter are shown in Figures~\ref{fig:IW_app}.
The sample-efficiency is improved by using the importance weight, especially on the HopperVel task.
Based on the results shown in Figure~\ref{fig:IW_app}, we used $c_{\textrm{clip}} = 0.3$ for LTD3 in our experiments.

\subsection{The frequency of minimizing the information loss}
\begin{figure}
	\centering
	\includegraphics[width=\columnwidth]{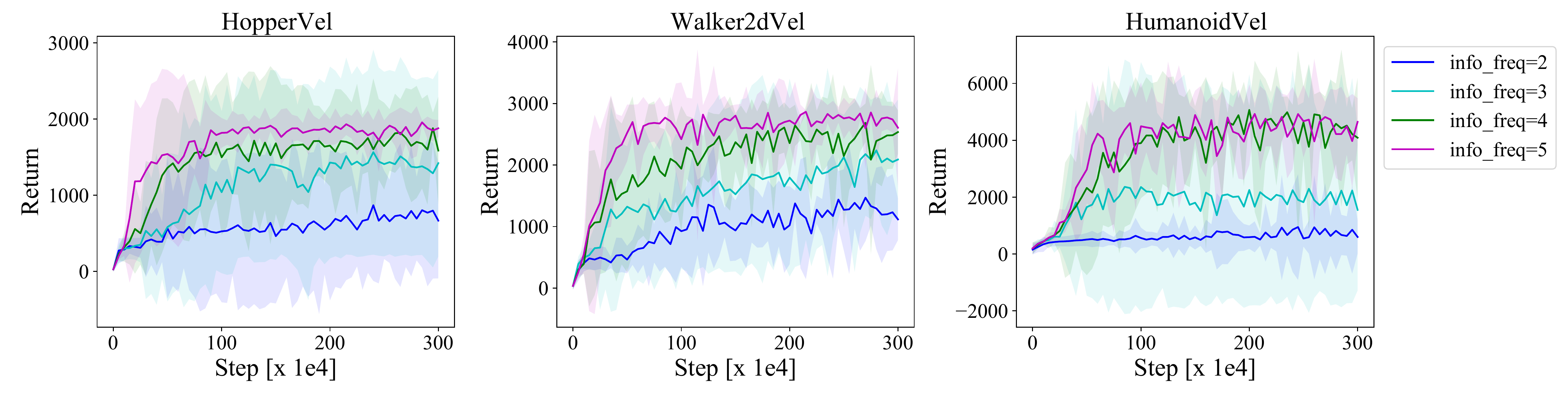}
	\caption{Effect of $d_{\textrm{info}}$ in LTD3. Results with two-dimensional continuous latent variable. Clipping parameter for the importance weight was $c_{\textrm{clip}} = 0.3$.}
	\label{fig:info_freq}
\end{figure}

The learning curves of LTD3 with different values of $d_{\textrm{info}}$ are shown in Figure~\ref{fig:info_freq}.
We observed a trade-off between the average return and the diversity of the solutions. 
If the information loss $\mathcal{J}_{\textrm{info}}$ is updated more often, then more diverse behaviors are obtained, but the average return decreases.
Based on the results shown in Figure~\ref{fig:info_freq}, we used $d_{\textrm{info}} = 4$ for LTD3 in our experiments.

\subsection{Learning Curve with Different Dimensionality of Latent Variable}
\begin{figure*}[t]
	%	\vspace{-1cm}
	\centering
	\includegraphics[width=\columnwidth]{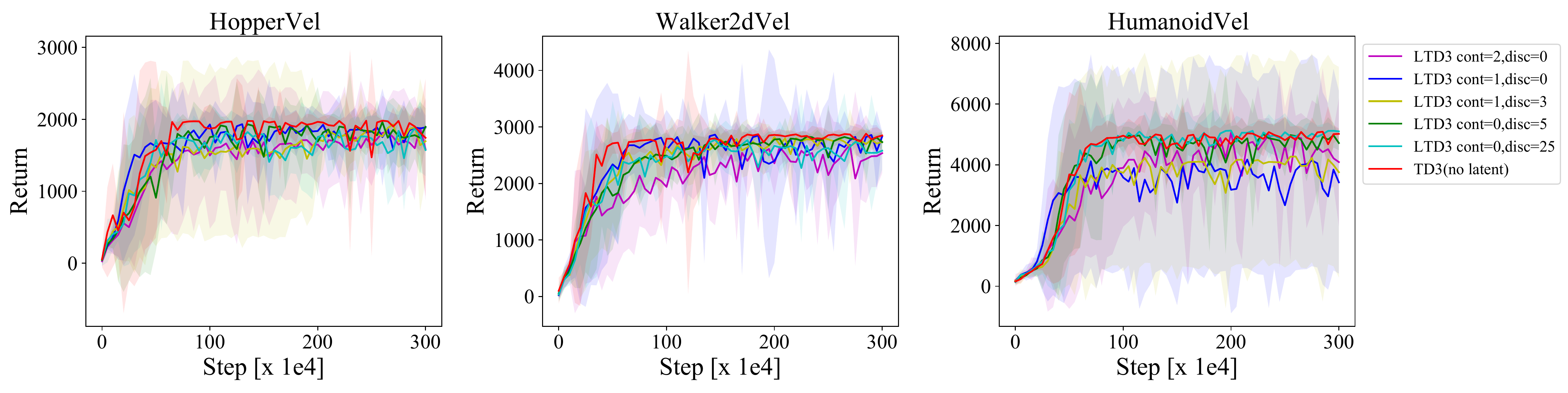}
	%	\vspace{-0.2cm}
	\caption{Learning curve of LTD3 with different dimensionalities of latent variable. }
	\label{fig:ltd3_dim}
	\vspace{-0.4cm}
\end{figure*}

The learning curves of LTD3 with different dimensionalities of the latent variable are shown in Figure~\ref{fig:ltd3_dim}. 
In LTD3, increasing the dimensionality of the latent variable did not necessarily increase the sample complexity in the training phase.
In the HumanoidVel task, the variance of the return of the policy with the two-dimensional latent variable was lower than that of the policy with the one-dimensional latent variable.

\section{Experimental Details}
\label{app:exp}
When plotting the learning curve, we performed the experiments five times with different random seeds, and reported the averaged test return where the test return was computed once every 5000 time steps by executing 10 episodes without exploration.

In few-shot adaptation tasks, we trained a latent-conditioned policy on a training MDP five times with different random seeds.
In a test MDP, five policies trained with different random seeds were adopted. For each random seed, the value of the latent value that yielded the best performance on the test MDP was determined after $k$ episodes in the test MDP; subsequently, the performance of the selected policy was evaluated in five episodes. 

\subsection{Details of Training MDPs}
In the training MDPs, we used the modified version of MuJoCo tasks implemented in OpenAI Gym.
The reward functions of the MuJoCo tasks were modified based on the study in \cite{Kumar20}.

\paragraph{HopperVel task}
The reward function of the original Hopper task in OpenAI Gym is given by
\begin{align}
	r = 	r_{\textrm{ctrl}} + 	r_{\textrm{vel}} + 	r_{\textrm{survive}},
\end{align}
where $r_{\textrm{ctrl}}$ is the control cost defined as
\begin{align}
	r_{\textrm{ctrl}} = - 1e^{-3} \left\| \vect{a} \right\|^2_2
\end{align}
$r_{\textrm{vel}}$ is the term for encouraging the agent to walk faster and is given by
\begin{align}
	r_{\textrm{vel}} = ( x_{t} - x_{t-1})) / \Delta t
\end{align}
$r_{\textrm{survive}}$ is the term for avoiding falling down and is defined as
\begin{align}
	r_{\textrm{survive}} =
	\left\{
	\begin{array}{cl}
		1 & \textrm{if} \ \vect{s}_t \ \textrm{is not terminal}\\
		0 & \textrm{otherwise}
	\end{array}
	\right. .
\end{align}

For the HopperVel task, we modified $r_{\textrm{vel}}$ as follows:
\begin{align}
	r_{\textrm{vel}} = \min( (x_{t} - x_{t-1})/ \Delta t , 1).
\end{align}
Other terms were used as in the original task.

\paragraph{Walker2dVel task}
The reward function of the original Walker2d task in OpenAI Gym is given by
\begin{align}
	r = 	r_{\textrm{ctrl}} + 	r_{\textrm{vel}} + 	r_{\textrm{survive}},
\end{align}
where $r_{\textrm{ctrl}}$ is the control cost given by
\begin{align}
	r_{\textrm{ctrl}} = - 1e^{-3} \left\| \vect{a} \right\|^2_2
\end{align}
$r_{\textrm{vel}}$ is the term for encouraging the agent to walk faster and is defined as
\begin{align}
	r_{\textrm{vel}} = ( x_{t} - x_{t-1})) / \Delta t
\end{align}
$r_{\textrm{survive}}$ is the term for avoiding falling down and is defined as
\begin{align}
	r_{\textrm{survive}} =
	\left\{
	\begin{array}{cl}
		1 & \textrm{if} \ \vect{s}_t \ \textrm{is not terminal}\\
		0 & \textrm{otherwise}
	\end{array}
	\right. .
\end{align}

For the Walker2dVel task, we modified $r_{\textrm{vel}}$ as follows:
\begin{align}
	r_{\textrm{vel}} = \min( (x_{t} - x_{t-1}) / \Delta t , 2).
\end{align}
Other terms were used as in the original task.

\paragraph{HumanoidVel task}
The reward function of the original Humanoid task in OpenAI Gym is given by
\begin{align}
	r = 	r_{\textrm{ctrl}} + r_{\textrm{vel}} + 	r_{\textrm{survive}} + r_{\textrm{impact}},
\end{align}
where $r_{\textrm{ctrl}}$ is the control cost defined as
\begin{align}
	r_{\textrm{ctrl}} = - 0.1  \left\| \vect{a} \right\|^2_2
\end{align}
$r_{\textrm{vel}}$ is the term for encouraging the agent to walk faster and is given by
\begin{align}
	r_{\textrm{vel}} = 0.25 ( x_{t} - x_{t-1})) / \Delta t
\end{align}
$r_{\textrm{survive}}$ is the term for avoiding falling down and is defined as
\begin{align}
	r_{\textrm{survive}} =
	\left\{
	\begin{array}{cl}
		5 & \textrm{if} \ \vect{s}_t \ \textrm{is not terminal}\\
		0 & \textrm{otherwise}
	\end{array}
	\right. 
\end{align}
$r_{\textrm{impact}}$ is given by
\begin{align}
	r_{\textrm{impact}} = - \min( 5e^{-6}\left\| \vect{f}_{\textrm{ext}} \right\|^2_2 , 10),
\end{align}
$\vect{f}_{\textrm{ext}}$ is the force exerted by the floor.

For the HumanoidVel task, we modified $r_{\textrm{vel}}$ to 
\begin{align}
	r_{\textrm{vel}} = 0.25  \min( (x_{t} - x_{t-1}) / \Delta t , 4).
\end{align}
Other terms were used as in the original task.

\paragraph{HalfCheetahVel task}
The reward function of the original HalfCheetah task in OpenAI Gym is given by
\begin{align}
	r = 	r_{\textrm{ctrl}} + 	r_{\textrm{vel}},
\end{align}
where $r_{\textrm{ctrl}}$ is the control cost given by
\begin{align}
	r_{\textrm{ctrl}} = - 0.1 \cdot \left\| \vect{a} \right\|^2_2
\end{align}
$r_{\textrm{vel}}$ is the term for encouraging the agent to walk faster and is defined as
\begin{align}
	r_{\textrm{vel}} = ( x_{t} - x_{t-1})) / \Delta t.
\end{align}

For the HalfCheetahVel task, we modified $r_{\textrm{vel}}$ as follows:
\begin{align}
	r_{\textrm{vel}} = \min( (x_{t} - x_{t-1}) / \Delta t , 1).
\end{align}
In addition, we did not use the control cost term. 

\paragraph{AntVel task}
The reward function of the original Ant task in OpenAI Gym is given by
\begin{align}
		r = 	r_{\textrm{ctrl}} + r_{\textrm{vel}} + 	r_{\textrm{survive}} + r_{\textrm{contact}},
\end{align}
where $r_{\textrm{ctrl}}$ is the control cost given by
\begin{align}
	r_{\textrm{ctrl}} = - 0.1 \cdot \left\| \vect{a} \right\|^2_2,
\end{align}
and $r_{\textrm{vel}}$ is the term for encouraging the agent to walk faster and is defined as
\begin{align}
	r_{\textrm{vel}} = ( x_{t} - x_{t-1})) / \Delta t.
\end{align}
The third term $r_{\textrm{survive}}$ is the term for avoiding falling down and is defined as
\begin{align}
	r_{\textrm{survive}} =
	\left\{
	\begin{array}{cl}
		1 & \textrm{if} \ \vect{s}_t \ \textrm{is not terminal}\\
		0 & \textrm{otherwise}
	\end{array}
	\right. 
\end{align}
The forth term $r_{\textrm{contact}}$ is given by
\begin{align}
	r_{\textrm{contact}} = - 0.005 \cdot \left\| \vect{f}_{\textrm{ext}} \right\|_2,
\end{align}
where $\vect{f}_{\textrm{ext}}$ is the force exerted by the floor.

For the AntVel task, we modified $r_{\textrm{vel}}$ as follows:
\begin{align}
	r_{\textrm{vel}} = \min( (x_{t} - x_{t-1}) / \Delta t , 1).
\end{align}
Other terms were used as in the original task.

\subsection{Few-Shot Adaptation Tasks}
\label{app:fewshot}
In the few-shot adaptation experiment, we evaluated LTD3, SMERL and SAC-DIAYN with two-dimensional continuous and discrete latent variables.
For the discrete latent variables, the dimension of the latent variable was set to $k$ as $|Z| = k$. Therefore, $k$ skills were learned on the training environment, and each learned skill was tested once in the test environment. 
For methods with continuous latent variables, the latent variables were uniformly sampled from $[-1, 1]$, and the performance of the policy with different values of the latent variable was tested in $k$ episodes. 
Subsequently, the value of the latent variable with the best performance was used to evaluate the performance after the few-shot adaptation.
For the few-shot adaptation task, the reward function for the test MDPs is the same as the one in the original reward function for the Walker2d task in OpenAI Gym.

\paragraph{W-Short1} The link under the knee joint of the red leg is shorter than that of the agent in the original Walker2d task in OpenAI Gym.

\paragraph{W-Short2} The link under the knee joint of the orange leg is shorter than that of the agent in the original Walker2d task in OpenAI Gym.

\paragraph{W-LowShort} The link under the knee joint of the red leg is shorter than that of the agent in the original Walker2d task in OpenAI Gym. In addition, the position of the knee joint of the orange leg is lower than that of the agent in the original Walker2d task in OpenAI Gym.

\paragraph{W-ShortHigh}  The link under the knee joint of the orange leg is shorter than that of the agent in the original Walker2d task in OpenAI Gym. In addition, the position of the knee joint of the red leg is higher than that of the agent in the original Walker2d task in OpenAI Gym.

\subsection{Hyperparameters for Training}
The hyperparameters of our methods and the baseline methods used in the experiments are summarized in Tables~\ref{tbl:hyperparam_td3}--\ref{tbl:hyperparam_ltd3}.
The hyperparameters used for SMERL and SAC-DIAYN were based on the study in \cite{Kumar20}.

\begin{figure}
	\begin{minipage}{0.45\textwidth}
		\begin{table}[H]
			\caption{Hyperparameters used for TD3}
			\label{tbl:hyperparam_td3}
			%			\vskip 0.15in
			\begin{center}
						\begin{tabular}{lc}
							\hline
							Parameter & Value  \\
							\hline
							Optimizer    & Adam \\
							Learning rate & $3 \cdot 10^{-4}$ \\
							Discount factor $\gamma$ & 0.99 \\
							Replay buffer size & $10^6$ \\
							Number of hidden layers & 2 \\
							Number of hidden units per layer & 256 \\
							Number of samples per minibatch & 256 \\
							Activation function & Relu \\
							Target smoothing coefficient & 0.005 \\
							gradient steps per time step & 1 \\
							interval for updating the policy & 2 \\
							\hline
						\end{tabular}
			\end{center}
			\vskip -0.1in
		\end{table}
	\end{minipage}
	\hfill
	\begin{minipage}{0.45\textwidth}
		\begin{table}[H]
			\caption{Hyperparameters used for SAC}
			\label{tbl:hyperparam_sac}
			%		\vskip 0.15in
			\begin{center}
						\begin{tabular}{lc}
							\hline
							Parameter & Value  \\
							\hline
							Optimizer    & Adam \\
							Learning rate & $3 \cdot 10^{-4}$ \\
							Discount factor $\gamma$ & 0.99 \\
							Replay buffer size & $10^6$ \\
							Number of hidden layers & 2 \\
							Number of hidden units per layer & 256 \\
							Number of samples per minibatch & 256 \\
							Activation function & Relu \\
							Target smoothing coefficient & 0.005 \\
							gradient steps per time step & 1 \\
							temperature & 0.2 \\
							\hline
						\end{tabular}
			\end{center}
			\vskip -0.1in
		\end{table}
	\end{minipage}
\end{figure}

\begin{figure}
	\begin{minipage}{0.45\textwidth}
		\begin{table}[H]
			\caption{Hyperparameters used for SAC-DIAYN}
			\label{tbl:hyperparam_sac_diayn}
			\begin{center}
						\begin{tabular}{lc}
							\hline
							Parameter & Value  \\
							\hline
							Optimizer    & Adam \\
							Learning rate & $3 \cdot 10^{-4}$ \\
							Discount factor $\gamma$ & 0.99 \\
							Replay buffer size & $10^6$ \\
							Number of hidden layers & 2 \\
							Number of hidden units per layer & 256 \\
							Number of samples per minibatch & 256 \\
							Activation function & Relu \\
							Target smoothing coefficient & 0.005 \\
							gradient steps per time step & 1 \\
							temperature & 0.1\\
							Weight on unsupervised reward & 0.5 \\
							\hline
						\end{tabular}
			\end{center}
			\vskip -0.1in
		\end{table}
	\end{minipage}
	\hfill
	\begin{minipage}{0.45\textwidth}
		\begin{table}[H]
			\caption{Hyperparameters used for SMERL}
			\label{tbl:hyperparam_smearl}
			\begin{center}
						\begin{tabular}{lc}
							\hline
							Parameter & Value  \\
							\hline
							Optimizer    & Adam \\
							Learning rate & $3 \cdot 10^{-4}$ \\
							Discount factor $\gamma$ & 0.99 \\
							Replay buffer size & $10^6$ \\
							Number of hidden layers & 2 \\
							Number of hidden units per layer & 256 \\
							Number of samples per minibatch & 256 \\
							Activation function & Relu \\
							Target smoothing coefficient & 0.005 \\
							gradient steps per time step & 1 \\
							temperature for SAC & 0.1 \\
							Weight on unsupervised reward & 10.0 \\
							Value of $\epsilon$ & 0.1 \\
							\hline
						\end{tabular}
			\end{center}
			\vskip -0.1in
		\end{table}
	\end{minipage}
\end{figure}

\begin{table}[tb]
	\caption{Hyperparameters used for LTD3}
	\label{tbl:hyperparam_ltd3}
	\begin{center}
				\begin{tabular}{lc}
					\hline
					Parameter & Value  \\
					\hline
					Optimizer    & Adam \\
					Learning rate & $3 \cdot 10^{-4}$ \\
					Discount factor $\gamma$ & 0.99 \\
					Replay buffer size & $10^6$ \\
					Number of hidden layers & 2 \\
					Number of hidden units per layer & 256 \\
					Number of samples per minibatch & 256 \\
					Activation function & Relu \\
					Target smoothing coefficient & 0.005 \\
					gradient steps per time step & 1 \\
					Clipping param. for IW & 0.3 \\
					\makecell[l]{interval for maximizing  $\mathcal{J}_{\textrm{Q}}(\vect{\theta})$} & 2 \\
					\makecell[l]{interval for maximizing $\mathcal{J}_{\textrm{info}}(\vect{\theta})$} & 4 \\
					\hline
				\end{tabular}
	\end{center}
	\vskip -0.1in
\end{table}

\bibliographystyle{elsarticle-num} 
\bibliography{ltd3}


%\begin{thebibliography}{00}


%\bibitem{}

%\end{thebibliography}
\end{document}